\documentclass[runningheads]{llncs}

\usepackage{eccv}

\usepackage{algorithm}
\usepackage{algorithmic}
\usepackage[utf8]{inputenc} 
\usepackage{booktabs}       
\usepackage{amsfonts}       
\usepackage{nicefrac}       
\usepackage{microtype}      
\usepackage{xcolor}         
\usepackage{dsfont}
\usepackage{multirow}
\usepackage{newfloat}
\usepackage{listings}
\usepackage{tikz}
\usepackage{eccvabbrv}
\usepackage{xcolor,colortbl}
\usepackage{graphicx}
\usepackage{booktabs}

\usepackage[accsupp]{axessibility}  

\usepackage{algorithm}
\usepackage[algo2e, ruled]{algorithm2e}
\SetKwComment{Comment}{// }{}
\SetKwProg{Init}{Initialize}{}{}

\makeatletter
\renewcommand{\fnum@figure}{Figure \thefigure}
\makeatother

\newcommand{\method}[1]{\textsc{Maximum Magnitude Selection}}
\newcommand{\methodshort}[1]{\textsc{MagMax}}

\usepackage{hyperref}

\usepackage{orcidlink}
\usepackage{wrapfig,booktabs}

\begin{document}

\title{\methodshort{}: Leveraging Model Merging\\for Seamless Continual Learning}


\author{
Daniel~Marczak\thanks{Corresponding author, email: \href{mailto:daniel.marczak.dokt@pw.edu.pl}{daniel.marczak.dokt@pw.edu.pl}}\inst{1,2}\orcidlink{0000-0002-6352-9134} \and
Bartłomiej~Twardowski\inst{1,5,6}\orcidlink{0000-0003-2117-8679} \and \\
Tomasz~Trzciński\inst{1,2,4}\orcidlink{0000-0002-1486-8906} \and
Sebastian~Cygert\inst{1,3}\orcidlink{0000-0002-4763-8381}
}

\authorrunning{D.~Marczak et al.}

\institute{
IDEAS~NCBR \and
Warsaw~University~of~Technology \and 
Gdańsk~University~of~Technology \and 
Tooploox \and
Autonomous~University of~Barcelona \and 
Computer~Vision~Center
}

\maketitle

\begin{abstract}

This paper introduces a continual learning approach named \methodshort{}, which utilizes model merging to enable large pre-trained models to continuously learn from new data without forgetting previously acquired knowledge. Distinct from traditional continual learning methods that aim to reduce forgetting during task training, \methodshort{} combines sequential fine-tuning with a maximum magnitude weight selection for effective knowledge integration across tasks. Our initial contribution is an extensive examination of model merging techniques, revealing that simple approaches like weight averaging and random weight selection surprisingly hold up well in various continual learning contexts. More importantly, we present \methodshort{}, a novel model-merging strategy that enables continual learning of large pre-trained models for successive tasks. Our thorough evaluation demonstrates the superiority of \methodshort{} in various scenarios, including class- and domain-incremental learning settings. The code is available at \url{https://github.com/danielm1405/magmax}.
  
\end{abstract}

\section{Introduction}
\label{sec:intro}

\begin{figure}[t]
    \centering
    \includegraphics[width=0.95\linewidth]{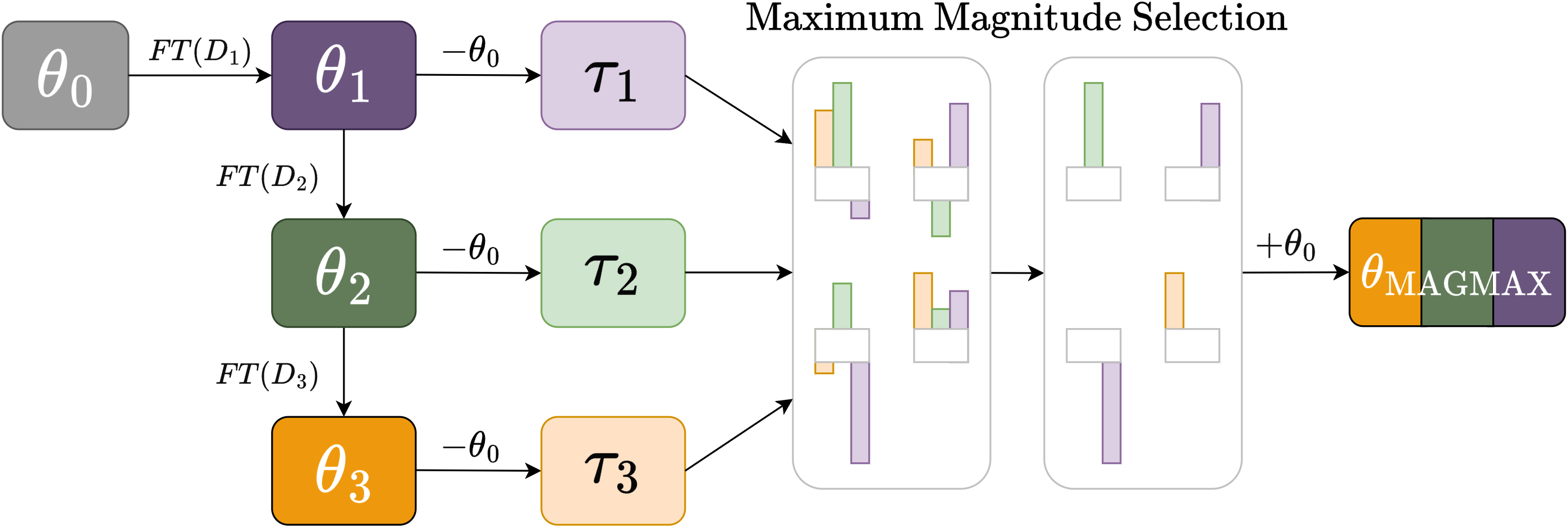}
    \caption{Overview of the proposed \methodshort{} method for continual learning. We sequentially fine-tune the model on the subsequent tasks and create task vectors $\tau_i$ by subtracting the weights of the pre-trained model $\theta_0$. Then we merge the task vectors using \method{} strategy which selects the parameters of task vectors by highest magnitude. Finally, we apply merged task vector to the pre-trained model to obtain a multitask model $\theta_{\methodshort{}}$. Note that with running statistics implementation we can only store two sets of weights (see Section~\ref{sec:mem_footprint} for details).}
    \label{fig:teaser}
\end{figure}

{Large pre-trained models are considered cornerstones of complex machine learning systems, allowing unprecedented performance improvements across many challenging tasks~\cite{radford2021learning, DINO, kirillov2023segment, carion2020endtoend, wang2022yolov7, zhai2023sigmoid}. Yet their remarkable ability to generalize to unseen conditions is intrinsically limited by the stationary character of their training data. To keep up with the ever-changing world, these models should adapt continuously and assimilate knowledge from the stream of new data, which is the objective of Continual Learning (CL)~\cite{ven2022three, lange2019continual, masana2023class}.}

Traditionally, CL approaches used regularization to retain the knowledge from previous tasks~\cite{LwF, kirkpatrick2017overcoming}, grow the network while learning new tasks~\cite{yan2021dynamically, Progressive}, or use a replay buffer to limit the catastrophic forgetting~\cite{hou2019_lucir, Foster, zhao2020_maintaining}.
In this work, we argue that in the era of machine learning systems built on top of large pre-trained models, 
utilizing this foundation seems to present a more intuitive and effective strategy for continuous learning.
Model merging is a new paradigm of adapting pre-trained models. It allows to consolidate the knowledge of multiple independently fine-tuned task-specific models into one multi-task model without any additional training. There are various methods that base on selecting or interpolating the weights of task-specific models~\cite{yadav2023tiesmerging, ilharco2023task, ortizjimenez2023tangent, singh2020fusion, matena2021merging}.
Contrary to the traditional CL methods, which focus on alleviating forgetting \textit{during} training on new tasks, 
model merging allows to seamlessly consolidate the knowledge \textit{after} the training on new tasks leaving the training procedure unchanged.

When evaluated across a single, fixed set of diversified heterogeneous tasks~\cite{yadav2023tiesmerging, ilharco2023task, ortizjimenez2023tangent}, such as recognition of hand-written digits~\cite{lecun1998mnist}, satellite images~\cite{eurosat} or car models~\cite{cars}, model merging methods perform well. However, this evaluation benchmark is far from a realistic use case. Furthermore, it does not include real-life applications with the data coming from similar (but disjoint) distributions, \textit{e.g.} various kinds of medical imagery. Here, we fill this gap and extensively evaluate model merging techniques with different levels of task similarity (including class- and domain-incremental scenarios), varying number of tasks, and their granularity.
We find that the simplest merging baselines - weight averaging and random weight selection - work surprisingly well, often outperforming sophisticated merging strategies and CL approaches.

Our evaluation highlights a significant drawback of the existing methods. They fine-tune pre-trained models independently for each task foregoing the potential of knowledge transfer. To address this significant limitation, we propose \methodshort{}, a novel method for continual learning that utilizes sequential fine-tuning and model merging via maximum magnitude selection (see Figure~\ref{fig:teaser}). We show that sequential fine-tuning simplifies model merging by reducing the number of sign conflicts -- a major source of interference when merging models~\cite{yadav2023tiesmerging} -- between task-specific models while maximum magnitude selection chooses the important parameter values.
We investigate the effectiveness of the parameter selection strategy and examine the contribution of task vectors. Finally, we highlight a broader impact of our findings showing that merging via maximum magnitude selection can improve existing CL methods and that sequential fine-tuning improves the performance of models combined using various merging techniques.

To sum up, our contributions are as follows:
\begin{itemize}
    \item We identify and fill the gaps in model merging evaluation by benchmarking the existing merging strategies in diverse settings with tasks containing different classes or domains when varying the number of tasks, task similarity, and their granularity.
    \item We find that simple baselines -- weight averaging and random weight selection -- are very strong and often outperform the existing merging strategies.
    \item We propose \methodshort{}, a novel method for continual learning that sequentially fine-tunes the model and consolidates the knowledge by merging weights of task-specific models using maximum magnitude selection. \methodshort{} achieves state-of-the-art results on multiple continual learning benchmarks.
    \item We highlight the broader implications of our results demonstrating that merging with maximum magnitude selection in model merging enhances existing continual learning methods and that sequential fine-tuning facilitates other existing merging techniques.
\end{itemize}

\section{Related Work}

\noindent\textbf{Continual learning (CL)} is a setting where models learn a sequence of tasks with access to the data from the current task only. The goal is to achieve high performance on all the tasks from the sequence, with catastrophic forgetting of knowledge learned from the previous tasks being the main challenge~\cite{french1999catastrophic, Mccloskey89}. One prominent example of CL approaches are the  regularization-based methods. In EWC~\cite{kirkpatrick2017overcoming}, the authors propose to use the Fisher information matrix to estimate model weight importance (for previous tasks) which is then used to penalize changes of important model weight. On the other hand, regularization can be applied on the data level, \eg LwF~\cite{LwF} or DER~\cite{yan2021dynamically} penalizes changes in model predictions or features. Other CL approaches include adding more parameters as the number of tasks increases~\cite{YoonYLH18, Progressive}, or using memory buffer~\cite{yan2021dynamically, hou2019_lucir, Foster, zhao2020_maintaining} for data from old tasks, which is often undesirable due to the privacy concerns. 
In general, it seems that the best results are obtained by CL methods that favor stability, that is the model does not change much between consecutive learning tasks~\cite{Kim_2023_CVPR, rypesc2024divide}.
As a result, a plethora of methods were developed for CL scenarios which assumed large first task~\cite{petit2023fetril, zhu2022self}, or Large Pre-trained Model (LPM).

\noindent\textbf{Continual Learning of LPMs} became popular as capabilities (e.g., zero shot or out-of-distribution (OOD) performance) of foundation models became apparent~\cite{DINO,radford2021learning, kirillov2023segment, carion2020endtoend, wang2022yolov7, zhai2023sigmoid}.
A recent study questioned the utility of some CL methods, showing that by using a frozen model and nearest mean classifier can obtain competitive results~\cite{janson2022simple}.
Further advancements to the use of LPM were driven by using the prompting techniques~\cite{L0SRSPDP22L2p,SmithKGCKAPFK23CODA}. 
Alternatively, SLCA proposed a simple model that fine-tunes only the classification layer with a small learning rate~\cite{ZhangWKCW23}. In general, when using LPMs the focus in CL shifts towards maximal stability. 

\noindent\textbf{Weights interpolation} has recently emerged as an efficient technique for transfer learning that reduces forgetting. After fine-tuning LPM on target data, its weights are interpolated with the weights of (unchanged) LPM, which allows finding a good balance between accuracy on the target domain and zero-shot capabilities of LPM~\cite{wortsman2022robust}. Such an approach was further extended when merging models across multiple models for OOD performance~\cite{wortsman2022model} or in multi-task learning (i.e., Task Vectors~\cite{ilharco2023task}). Since then multiple methods have been developed in this area. TIES-Merging~\cite{yadav2023tiesmerging} reduces the interference when merging models by trimming parameters and electing signs. In~\cite{ortizjimenez2023tangent}, the authors linearize the fine-tuning to disentangle weights and facilitate merging. ZipLoRA~\cite{shah2023ZipLoRA} adapts diffusion models by merging LoRA weights for different styles and subjects. However, those methods were, up-to-date, evaluated on a limited number of scenarios.
In this work, we are interested in using those promising approaches to test how they work for different similarities between tasks, as well as when they are compared with simple CL baselines.
A concurrent work, CoFiMA~\cite{marouf2023weighted}, utilizes Fisher Merging~\cite{matena2021merging} sequentially after each task to continually train closed vocabulary image classifiers. In contrast, we focus on reducing parameter-level interferences in open vocabulaty models.

\section{Background and motivation}

\subsection{Problem setting}

We consider a problem of continual learning of large pre-trained models. We assume access to a pre-trained model parametrized by $d$ weights $\theta_0 \in \mathbb{R}^d$. Our goal is to adapt the model to a sequence of disjoint tasks $\{D_1, D_2, \dots, D_n\}$ one task at a time. We investigate \textit{exemplar-free} scenario which assumes no access to data from previous tasks.

We consider two fine-tuning scenarios:
\begin{itemize}
    \item independent (Ind FT) - starts from pre-trained weights $\theta_0$,
    \item sequential (Seq FT) - starts from the weights of the model fine-tuned on the sequence of previous tasks, \ie when fine-tuning on task $D_t$, we start from $\theta_{t-1}$ which was trained on $\{D_1, D_2, \dots, D_{t-1}\}$.
\end{itemize}

We use a notion of task vector~\cite{ilharco2023task} that is an element-wise difference between the fine-tuned model and the pre-trained model, \ie $\tau_i = \theta_i - \theta_0$.
Note that independently fine-tuned task vectors contain information about a single task and sequentially fine-tuned task vectors encompass some knowledge about all the tasks in the sequence.

\subsection{Motivation}

In this Section, we set and experimentally validate two hypotheses that serve as a motivation for developing a new method for continual learning via model merging.

\begin{figure}[t]
  \begin{minipage}{.54\linewidth}
    \centering
    \includegraphics[width=1.04\linewidth]{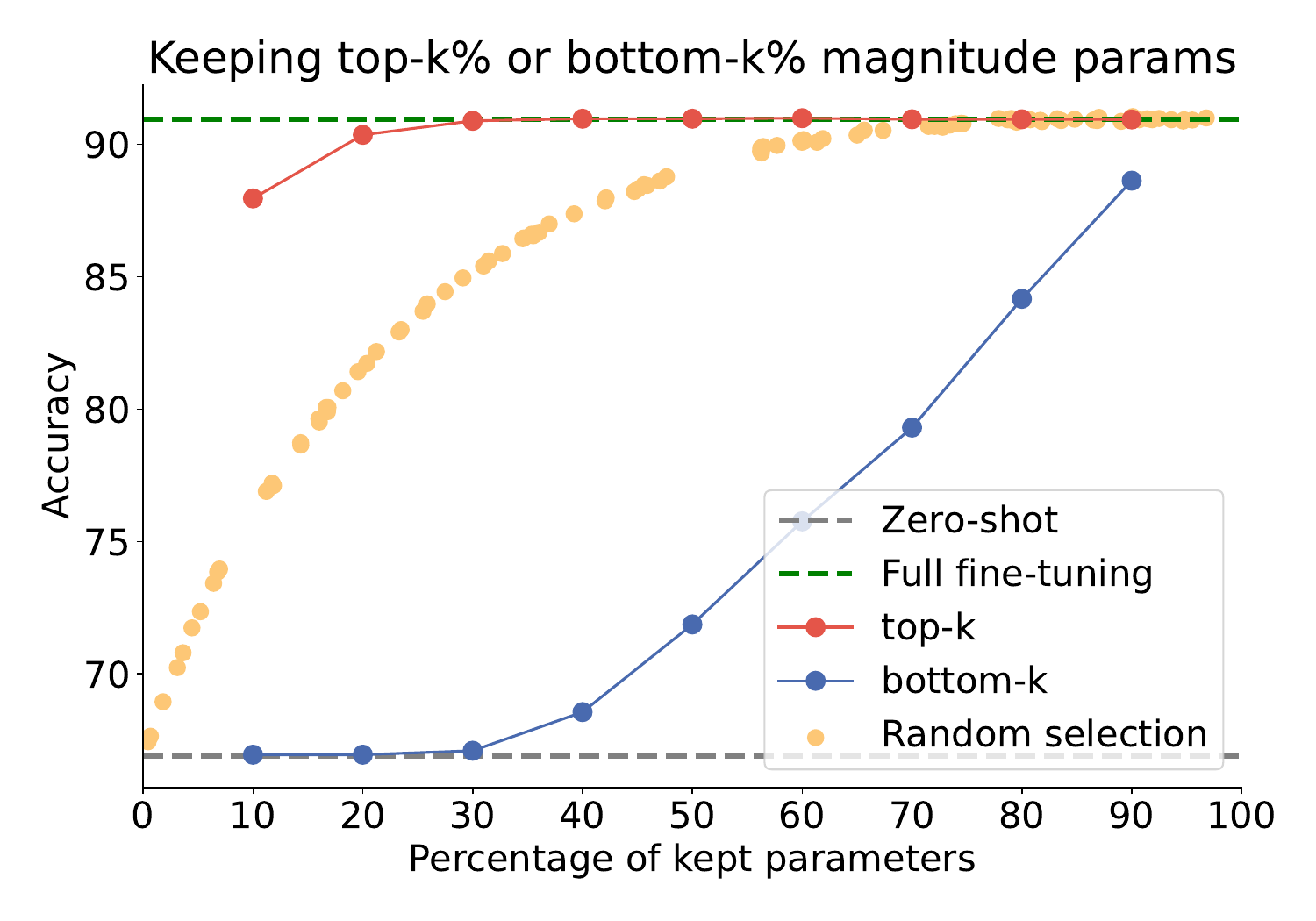}
    \caption{Only a small fraction of parameters that changed the most during fine-tuning is responsible for improved performance.}
    \label{fig:motivation-pruning}
  \end{minipage}
  \hfill
  \begin{minipage}{.44\linewidth}
    \centering
    \includegraphics[width=0.99\linewidth]{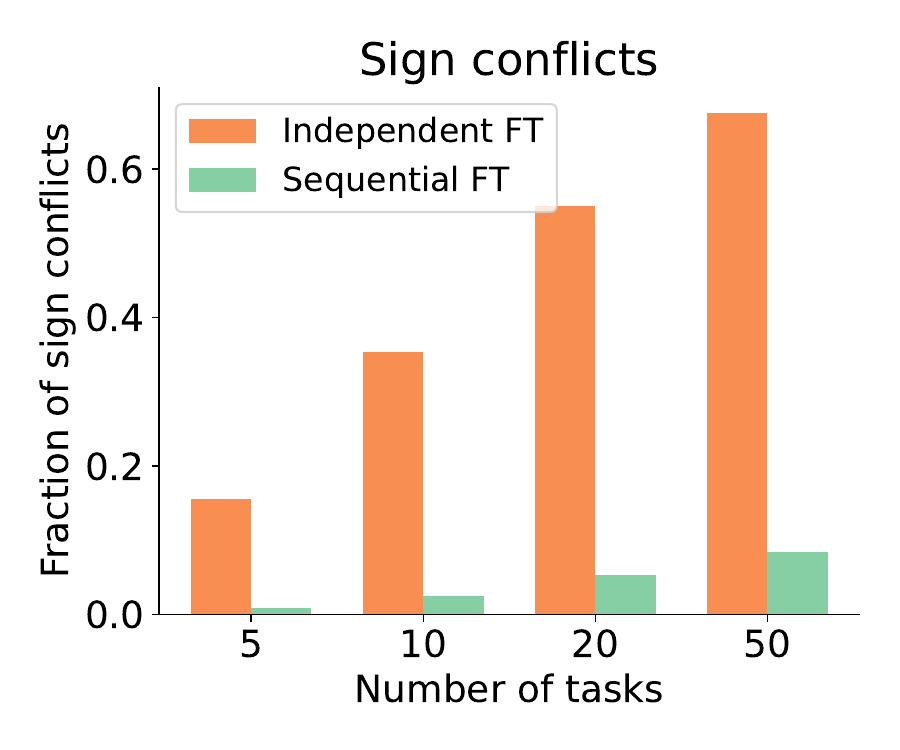}
    \caption{Sequential fine-tuning encourages consistent directions of parameter updates. We report sign conflicts after trimming 80\% of the lowest magnitude parameters in each task vector.}
    \label{fig:motivation-sign-conflicts}
  \end{minipage}
\end{figure}

\paragraph{\textbf{$\mathcal{H}1$: Parameters that change the most during fine-tuning are the most important for the task.}}

To verify this hypothesis we conduct the following experiment. We fine-tune a model on CIFAR100 dataset and create a task vector $\tau$. Then, we keep only $k\%$ of parameters that are selected at random, or according to their magnitude (lowest or highest) and remove the rest. Finally, we apply the pruned task vector to the pre-trained model and evaluate its performance\footnote{Note, that this experiment considers pruning parameters of task vector instead of pruning the weights of the network. Therefore, the conclusions may differ from neural pruning literature that considers magnitude pruning a strong baseline~\cite{han2015learning, guo2016dynamic, frankle2020linear}.}. Figure~\ref{fig:motivation-pruning} presents the results of this experiment. We observe that only a small fraction of high-magnitude parameters in task vectors are relevant for the model performance. Keeping only 20\% of the highest magnitude parameters yields results similar to fully fine-tuned models. To achieve similar performance we need to keep more than 90\% of the lowest magnitude parameters or more than 60\% of randomly selected parameters. These results validate $\mathcal{H}1$.

\paragraph{\textbf{$\mathcal{H}2$: Sequential fine-tuning reduces sign conflicts.}}
When fine-tuning the model on several tasks, sometimes we can observe a disagreement between the directions of task-specific updates. Such a situation is denoted as \emph{sign conflict}, as different task vectors have inconsistent signs for the same parameters. As noticed in~\cite{yadav2023tiesmerging} merging models with sign conflicts results in interference between tasks, and hence reduced performance of the final model. In this work, we postulate that sequential fine-tuning can reduce the number of sign conflicts. 
To verify this hypothesis, we fine-tune a model on CIFAR100 split into various number of tasks and count the conflicts of top-20\% parameters in corresponding task vectors. We perform fine-tuning either independently or sequentially.
Figure~\ref{fig:motivation-sign-conflicts} presents the results. We observe that sequential fine-tuning significantly reduces the sign conflicts validating $\mathcal{H}2$.

\section{\method{}}

Based on the motivations introduced in the previous Section, we introduce \method{} (\methodshort{}). It is a novel method for continual learning that utilizes sequential fine-tuning, following $\mathcal{H}2$, and model merging based on selecting the parameters of the highest magnitude, following $\mathcal{H}1$ (see Algorithm~\ref{alg:method}).
Given a new task, $D_t$, our method consists of two steps:
\begin{enumerate}

    \item \textbf{Sequential adaptation:} We obtain the new weights of the model $\theta_t$ by fine-tuning it on $D_t$. Importantly, we start from the weights of the model fine-tuned on previous tasks $\theta_{t-1}$.

    \item \textbf{Knowledge consolidation:} We consolidate task-specific knowledge using model merging. Firstly, we create task vectors for all tasks seen so far: $\{\tau_i\}_{i=1}^{t}$, where $\tau_i = \theta_i - \theta_0$. Then, for each parameter $p \in \{1,2, \hdots, d\}$, we select the value $\tau_{\methodshort{}}^p$ by the maximum magnitude out of all the task vectors. Lastly, we apply the resulting task vector $\tau_{\methodshort{}}$ to the pre-trained model $\theta_{\methodshort{}} = \theta_0 + \lambda * \tau_{\methodshort{}}$, where $\lambda$ is a scaling factor.

\end{enumerate}

\begin{algorithm}[H]
\caption{\label{alg:method} Continual learning with \methodshort{}}
\DontPrintSemicolon
\KwIn{Pre-trained model $\theta_0$ with $d$ parameters, sequence of tasks $\{D_t\}_{t=1}^N$}
\For{$t$ $\textbf{in}$ $1,..., N$}
{
    $\theta_t \leftarrow \text{fine-tune}(\theta_{t-1}, D_t)$     \Comment{\scriptsize Fine-tune from previous checkpoint on current data}

    \For{$i$ $\textbf{in}$ $1,..., t$}
    {
        $\tau_i = \theta_i - \theta_0$     \Comment{\scriptsize Create task vectors}
    }
    \For{$p$ $\textbf{in}$ $1,..., d$}
    {
        $k \leftarrow \underset{i}{\arg\max} \{ | \tau^p_i | \}_{i=1}^t$
        
        $\tau_{\methodshort{}}^p \leftarrow \tau_k^p $     \Comment{\scriptsize Maximum Magnitude Selection}
    }
    $\theta_{\methodshort{}} \leftarrow \theta_0 + \lambda * \tau_{\methodshort{}}$    \Comment{\scriptsize Apply merged task vector to the pre-trained model}

    \Comment{\scriptsize Use model $\theta_{\methodshort{}}$ until new task}
}
\end{algorithm}

\section{Experimental setup}

\paragraph{\textbf{Datasets.}}

For class-incremental learning (CIL) experiments we use CIFAR100~\cite{krizhevsky2009learning} and ImageNet-R~\cite{hendrycks2020many} as generic image recognition benchmarks and CUB200~\cite{wah2011caltech} and Cars~\cite{cars} as fine-grained classification datasets. We split the datasets into $N$ equal subsets of disjoint classes, where $N \in \{5,10,20,50\}$ for generic benchmarks and $N \in \{5,10,20\}$ for fine-grained benchmarks (which contain less data).

To compare between class- and domain-incremental learning (DIL) we use ImageNet-R and DomainNet~\cite{peng2019moment}. For domain-incremental learning experiments, we split DomainNet into 6 tasks by their domain (clipart, infographics, painting, quickdraw, real and sketch) and ImageNet-R into 15 tasks by their renditions (including cartoons, origami, paintings, sculptures, \, etc). Moreover, we split these datasets into the corresponding number of tasks following class-incremental protocol (described in the previous paragraph) for a fair comparison of CIL and DIL performance.

We also study the eight task setup proposed by~\cite{ilharco2023task} that includes the following datasets: Cars~\cite{cars}, DTD~\cite{dtd}, SUN397~\cite{sun397}, EuroSAT~\cite{eurosat}, GTSRB~\cite{gtsrb}, MNIST~\cite{lecun1998mnist}, SVHN~\cite{svhn} and RESISC45~\cite{cheng2017remote}. This benchmark is widely popular in model merging community~\cite{yadav2023tiesmerging, ilharco2023task, ortizjimenez2023tangent}.

\paragraph{\textbf{Baselines.}}
We compare \methodshort{} against well-established CL baselines \textbf{LwF}~\cite{LwF} and \textbf{EWC}~\cite{kirkpatrick2017overcoming} as well as recent model merging strategies, Model Soup (\textbf{Avg})~\cite{wortsman2022model}, Task Arithmetic (\textbf{TA})~\cite{ilharco2023task} and TIES-Merging (\textbf{TIES})~\cite{yadav2023tiesmerging}. Additionally, we introduce a simple baseline dubbed \textbf{RandMix} which randomly selects each parameter from one of the fine-tuned models, \ie $\theta_m^p \sim \{\theta_i^p\}_{i=1}^N$. We also evaluate \textbf{MaxAbs} baseline, which is basically \methodshort{} with independent fine-tuning instead of sequential. Finally, we present \textbf{zero-shot} performance which denotes the capabilities of the pre-trained model, and \textbf{joint} performance of a model fine-tuned on the whole dataset.

\paragraph{\textbf{Implementation details.}}

We use CLIP pre-trained model~\cite{radford2021learning} with ViT/B-16~\cite{dosovitskiy2021an} image encoder. We follow the training procedure from~\cite{ilharco2022patching}, namely we fine-tune the image encoder with a batch size of 128, learning rate 1e-5, and a cosine annealing learning rate schedule and AdamW optimizer with weight decay 0.1. We train CIFAR100, ImageNet-R and DomainNet for 10 epochs each task, and CUB200 and Cars for 30 epochs. We use the final classification layer output by CLIP's text encoder and keep it frozen during fine-tuning, following~\cite{ilharco2022patching}. This fine-tuning recipe preserves the open-vocabulary nature of the model and does not harm the accuracy compared to training the classification layer~\cite{ilharco2022patching}.

We consider an exemplar-free continual learning scenario in which we cannot store any data from the previous tasks. As a result, we can not tune scaling factor $\lambda$ at merging time as described in~\cite{ilharco2023task}. Therefore, we follow no validation scenario from~\cite{yadav2023tiesmerging} and set constant $\lambda$ for each method based on experiments on CIFAR100/5 setting. We choose $\lambda=0.5$ for \methodshort{}, $\lambda=0.55$ for TIES and $\lambda=1/N$ for Task Vectors. Notice, that choosing $\lambda=1/N$ for Task Vectors simplifies the method to a simple average of task vectors. It makes Task Vectors and Model Soup identical, and we call this method Avg in further experiments. We tune the hyperparameters of CL methods in the same scenario, setting $\lambda=1e6$ for EWC and $\lambda=0.3$ for LwF.

\paragraph{\textbf{Memory footprint.}}
\label{sec:mem_footprint}

In Figure~\ref{fig:teaser} and Algorithm~\ref{alg:method}, we describe that \methodshort{} stores all the previous checkpoints for the sake of simplicity. However, an efficient implementation of the method stores two sets of weights: sequentially fine-tuned $\theta_t$ and combined task vector $\tau_{\methodshort{}_t}$ of running statistics (maximum magnitude). When task $t+1$ arrives, we start from $\theta_t$ and fine-tune the model resulting in $\theta_{t+1}$. Then, we merge $\tau_{\methodshort{}_t}$ with $\tau_{t+1}$ which is identical to merging $\{\tau_i \}_{i=0}^{t+1}$. That requires a constant memory footprint.

\section{Main results}

\paragraph{\textbf{Class-incremental learning.}}

Table~\ref{tab:results-CIL} presents the comparison of \methodshort{} with CL methods and merging-based baselines on various class-incremental learning benchmarks. \methodshort{} consistently outperforms the competitors across the scenarios that vary in number of tasks and dataset granularity, achieving on average 2.1\% better results than the second best method. Interestingly, simple baselines that merge independent fine-tunings by averaging (Avg) or \underline{even randomly mixing} (RandMix) the weights, are close competitors to CL methods and other merging strategies.

\begin{table}[b]
\centering
\caption{
\methodshort{} outperforms other continual learning methods and merging-based approaches on a wide variety of class-incremental scenarios. We report task-agnostic accuracy (\%) after the final task. The best results are in \textbf{bold} and the second best \underline{underlined}.
}
\scalebox{0.75}{
\begin{tabular}{l|cccc|cccc|ccc|ccc|c}
\toprule
 & \multicolumn{4}{c|}{CIFAR100} & \multicolumn{4}{c|}{ImageNet-R} & \multicolumn{3}{c|}{CUB200} & \multicolumn{3}{c|}{Cars} & \multicolumn{1}{c}{Avg}\\
Method & /5 & /10 & /20 & /50 & /5 & /10 & /20 & /50 & /5 & /10 & /20 & /5 & /10 & /20 & \\
\midrule
Zero-shot & \multicolumn{4}{c|}{66.91} & \multicolumn{4}{c|}{77.73} & \multicolumn{3}{c|}{56.08} & \multicolumn{3}{c|}{64.71} & 67.21 \\
Joint & \multicolumn{4}{c|}{90.94} & \multicolumn{4}{c|}{87.55} & \multicolumn{3}{c|}{81.57} & \multicolumn{3}{c|}{88.21} & 87.38 \\
\midrule
LwF & 83.25 & 73.45 & 72.05 & 68.84 & 81.15 & \underline{82.97} & \underline{81.82} & 80.32 & \textbf{65.12} & \underline{60.67} & \textbf{58.90} & \underline{71.72} & \textbf{69.84} & 62.98 & \underline{72.36} \\
EWC & \textbf{84.41} & 76.24 & \underline{75.39} & 72.97 & 82.15 & 82.42 & 81.48 & \underline{81.47} & 59.10 & 54.49 & 53.31 & 69.46 & 60.78 & 57.42 & 70.79 \\
RandMix & 81.55 & 77.04 & 75.36 & 72.91 & \underline{83.10} & 81.88 & 80.18 & 78.50 & 59.86 & 58.53 & \underline{58.08} & 67.32 & 65.62 & \underline{64.95} & 71.78 \\
MaxAbs & 81.95 & 76.75 & 74.39 & 73.04 & 83.03 & 82.33 & 80.92 & 79.33 & 60.15 & 58.01 & 56.59 & 67.36 & 63.55 & 58.95 & 71.17 \\
Avg & 81.41 & 77.04 & 75.29 & 72.92 & 83.08 & 81.87 & 80.27 & 78.53 & 59.77 & 58.44 & 58.01 & 67.37 & 65.59 & 64.88 & 71.75 \\
TIES & 81.72 & \underline{77.23} & 74.66 & \underline{73.76} & 83.08 & 82.27 & 80.83 & 79.57 & 60.94 & 58.22 & 56.97 & 70.45 & 64.90 & 61.17 & 71.84 \\
\midrule
\methodshort{} & \underline{84.16} & \textbf{80.41} & \textbf{78.49} & \textbf{76.75} & \textbf{83.60} & \textbf{83.33} & \textbf{82.27} & \textbf{81.75} & \underline{63.89} & \textbf{60.74} & \textbf{58.90} & \textbf{73.61} & \underline{69.28} & \textbf{65.84} & \textbf{74.50} \\
\bottomrule
\end{tabular}
}
\label{tab:results-CIL}
\end{table}

\paragraph{\textbf{Task-agnostic results.}}

\begin{figure}[h!]
    \centering
    \includegraphics[width=1.02\linewidth]{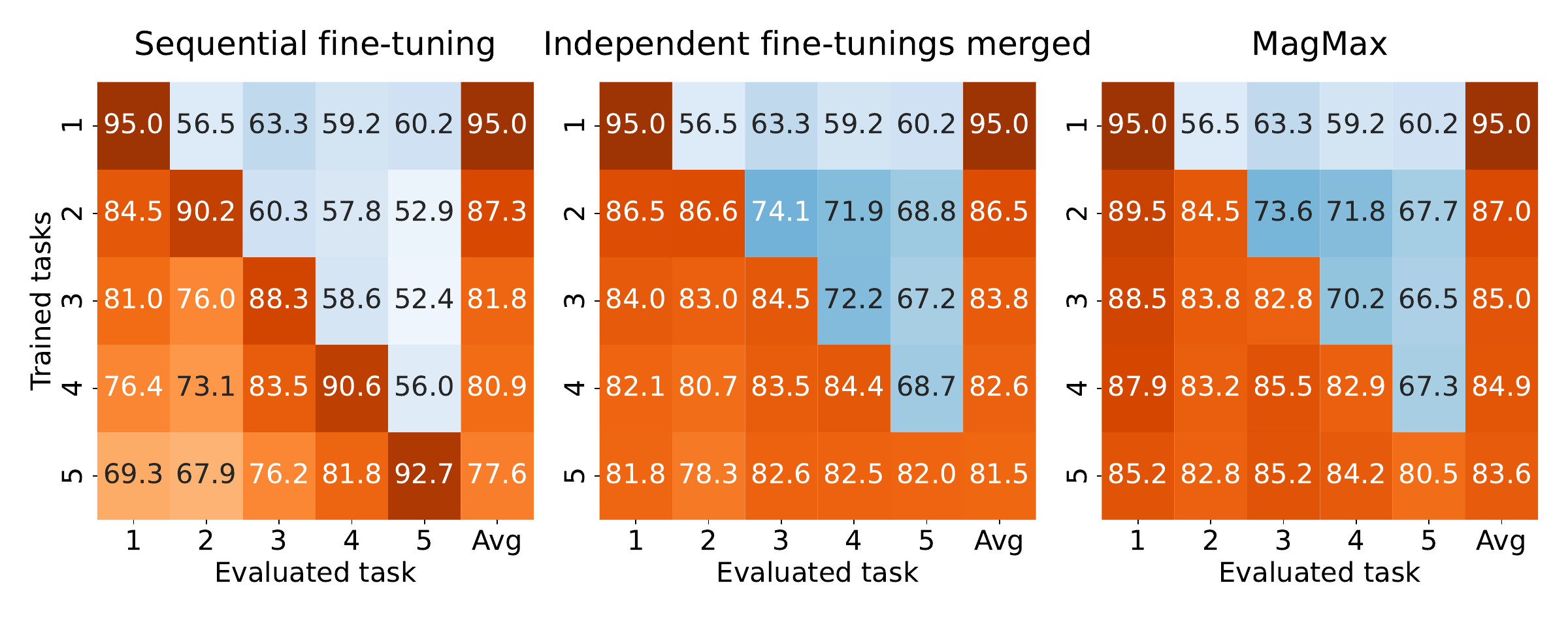}
    \caption{
        Sequential fine-tuning (left) exhibits high forgetting. Merging independent fine-tunings significantly reduces the forgetting (middle). \methodshort{} further improves this issue (right). We present the results on already learned tasks in orange and zero-shot performance in blue. We report task-agnostic accuracy (\%) for each task (columns) after training on the subsequent tasks (rows). The last column is an average accuracy on already seen tasks (lower triangular matrix in orange).
    }
    \label{fig:tag-results}
\end{figure}

Figure~\ref{fig:tag-results} presents task-agnostic results during continual learning for sequential fine-tuning, independent fine-tuning with model merging, and \methodshort{}. We observe that model merging significantly reduces forgetting: sequential fine-tuning exhibited 25.7\% forgetting on the first task and 22.3\% on the second one while \methodshort{} exhibited only 9.8\% and 1.7\%, respectively. Moreover, we observe significantly better performance on unseen tasks when using model merging.

\paragraph{\textbf{Domain-incremental learning.}}

\begin{figure}[h!]
\begin{minipage}{.44\linewidth}
    \centering
    \scalebox{0.92}{
    \begin{tabular}{l|cc|cc}
    \toprule
    & \multicolumn{2}{c|}{DomainNet} & \multicolumn{2}{c}{ImageNet-R} \\
    Method & CIL & DIL & CIL & DIL \\
    \midrule
    LwF & \textbf{69.59} & \underline{69.67} & \underline{82.05} & \underline{84.78} \\
    EWC & 64.73 & \textbf{70.74} & 81.45 & 83.77 \\
    RandMix & 65.60 & 64.31 & 80.85 & 82.28 \\ 
    MaxAbs & 66.21 & 67.51 & 81.50 & 83.93 \\
    Avg & 65.71 & 64.98 & 80.80 & 82.98 \\
    TIES & 66.62 & 66.42 & 81.52 & 83.90 \\
    \midrule
    \methodshort{} & \underline{69.18} & 69.00 & \textbf{82.90} & \textbf{85.40} \\
    \bottomrule
    \end{tabular}}
    \captionof{table}{
        \methodshort{} outperforms other merging-based methods in domain-incremental scenarios and achieves similar results to CL methods.
        We report task-agnostic accuracy (\%) after the final task. The best results are in \textbf{bold} and the second best \underline{underlined}.
    }
    \label{tab:results-CIL-vs-DIL}
\end{minipage}
\hfill
\begin{minipage}{.54\linewidth}
    \centering
    \includegraphics[width=1.03\linewidth]{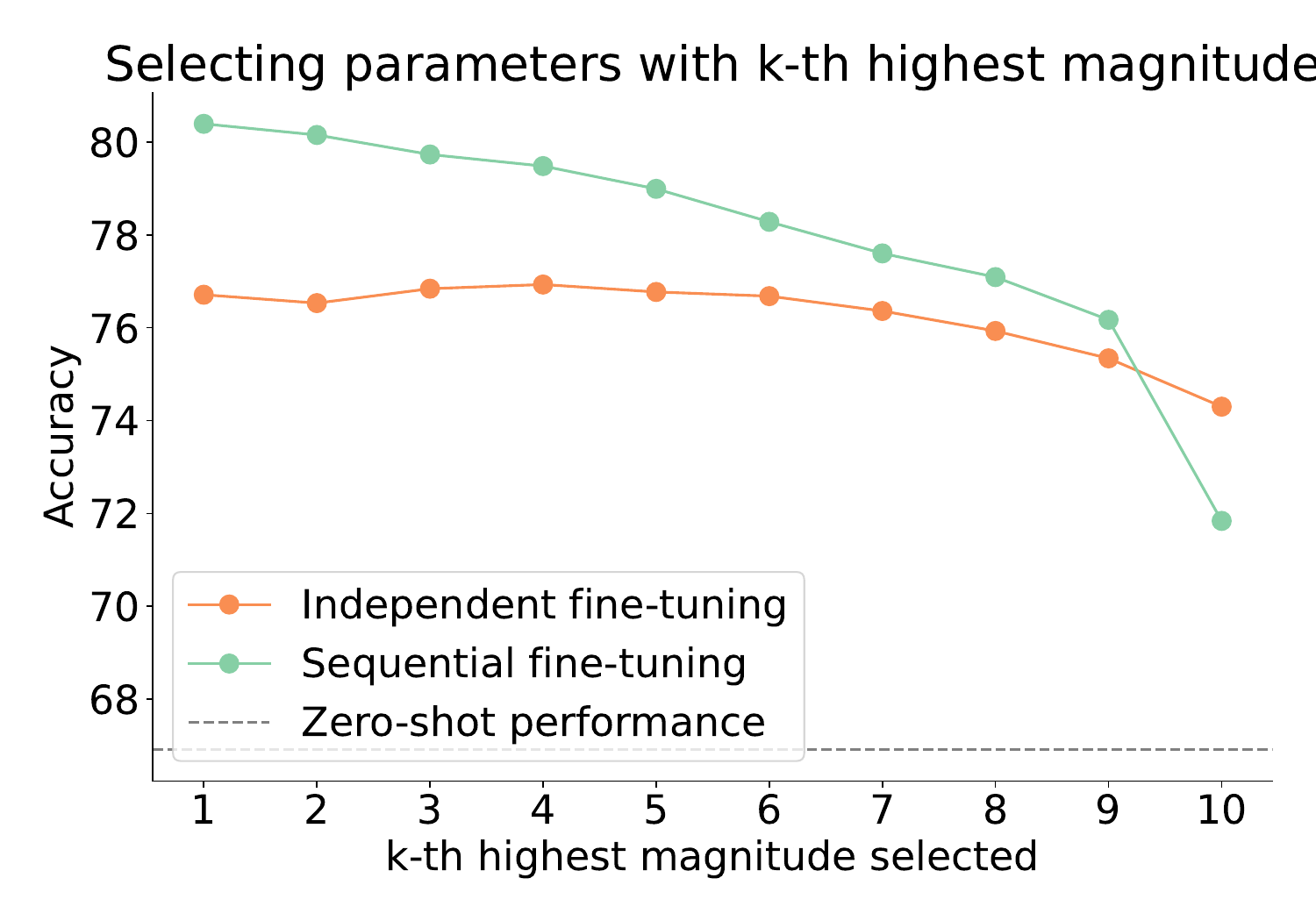}
    \caption{Selecting the highest magnitude parameters results in the best performance when merging sequentially fine-tuned models. We report the accuracy (\%) of the model merged by selecting $k$-th highest magnitude.}
    \label{fig:k-th-magnitude-merging}
\end{minipage}
\end{figure}

Table~\ref{tab:results-CIL-vs-DIL} presents the results on domain-incremental learning benchmarks. \methodshort{} outperforms other merging strategies in every scenario. It also achieves results on par with CL methods, outperforming them on ImageNet-R but slightly underperforming on DomainNet. We also observe that the top-performing methods achieve higher performance in domain-incremental scenarios than in class-incremental.

\paragraph{\textbf{Merging by $k$-th magnitude.}}

In this section, we experimentally justify the choice of maximum magnitude when merging models. We perform experiments where we merge task vectors by selecting the parameters that have $k$-th highest magnitude, where $k=1$ means maximum magnitude selection. We perform these evaluations for both independent and sequential fine-tuning scenarios. We also normalize the resulting task vectors so they have an equal norm for $k \in \{1, \dots, N\}$. We present the results in Figure~\ref{fig:k-th-magnitude-merging}. We observe that when fine-tuning independently, the results for $k \in \{1, \dots, 8\}$ vary by only 1\%. It means that the directions of updates defined by the resulting task vectors are similarly beneficial for the final performance. It suggests that parameters of independently fine-tuned models are either redundant (they serve the same purpose therefore the performance does not change) or concurrent (they serve concurrent task-specific purposes). However, for sequential fine-tuning, the performance decreases as $k$ increases. It means that parameters with high magnitude are better indicators of the beneficial update direction than parameters with lower magnitude.

\paragraph{\textbf{Selecting high magnitude parameters promotes consistent update directions.}}

\begin{figure}[t]
\begin{minipage}{.49\linewidth}
    \centering
    \includegraphics[width=1.0\linewidth]{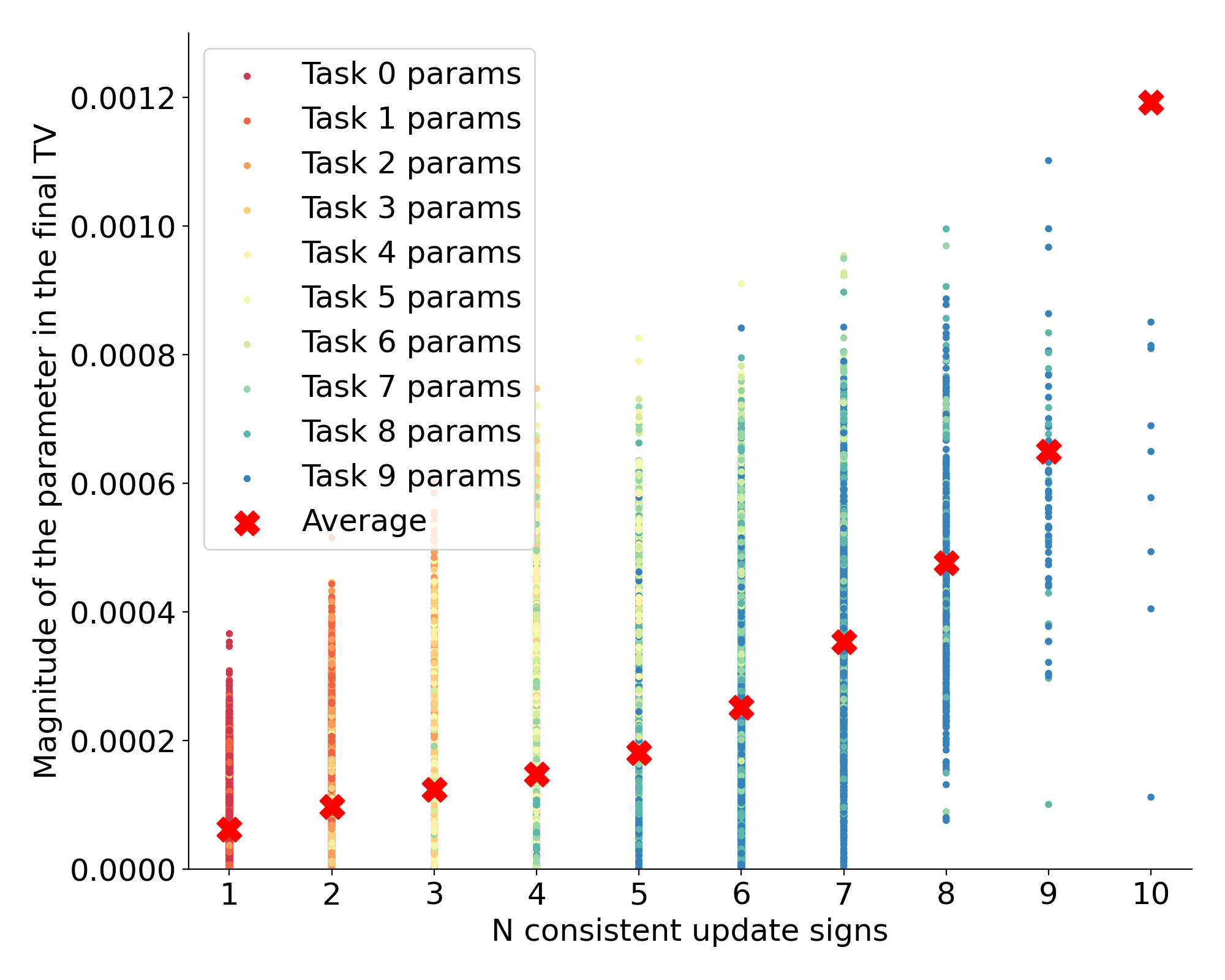}
    \caption{Magnitude of parameters of sequentially fine-tuned task vectors is correlated with the consistency of the update direction in the subsequent tasks. We report the results in CIFAR100/10 setting.}
    \label{fig:consistency-vs-magnitude}
\end{minipage}
\hfill
\begin{minipage}{.49\linewidth}
    \centering
    \includegraphics[width=1.0\linewidth]{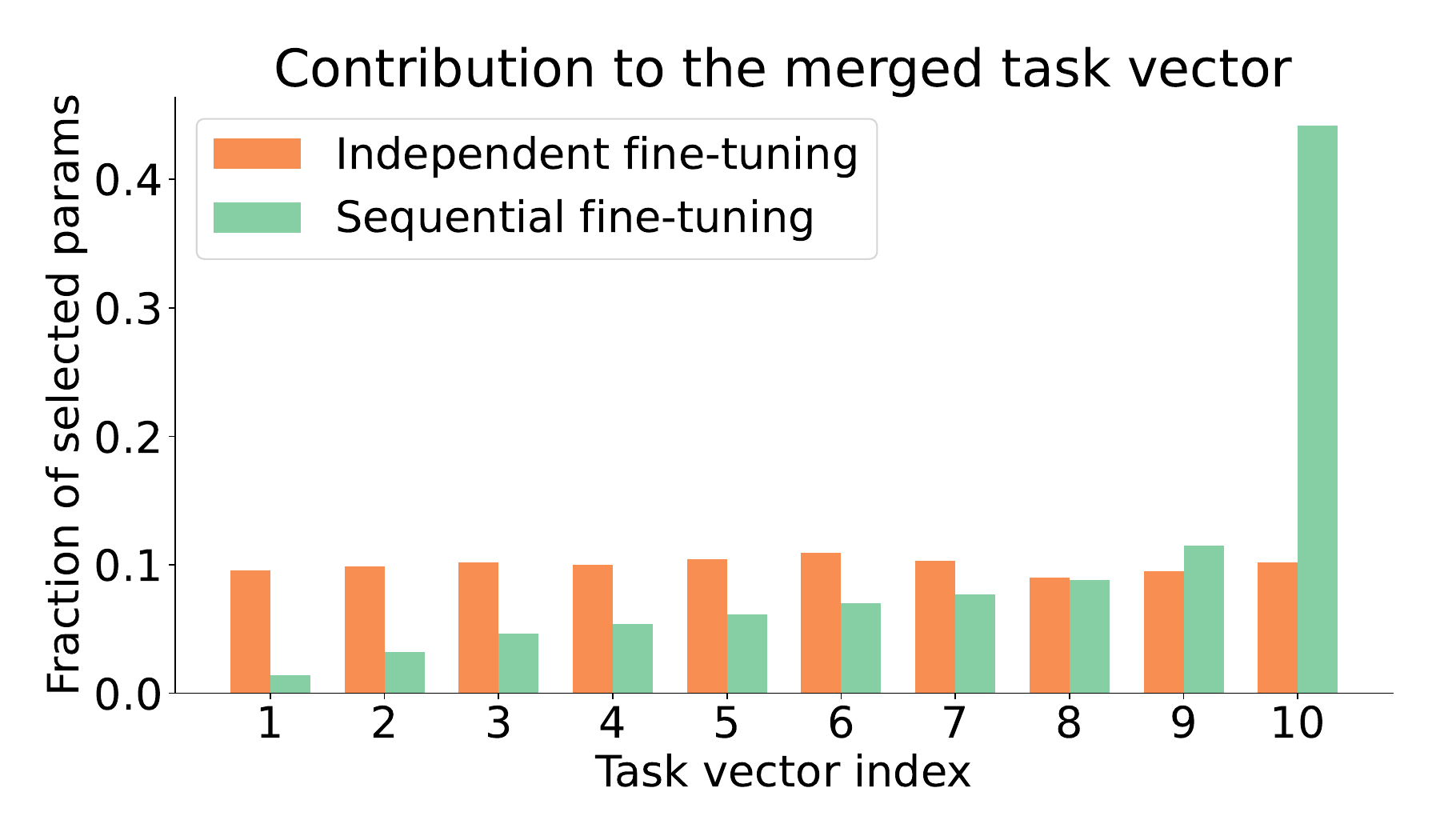}
    \caption{The contribution of parameters is nearly evenly distributed across task when fine-tuning independently. However, for sequential fine-tuning merging prioritizes the later task vectors which accumulated the knowledge about multiple tasks. We report the results in CIFAR100/10 setting.}
    \label{fig:task-vector-contribution}
\end{minipage}
\end{figure}

In this Section we set and verify the following hypothesis: \textit{parameters which update directions were consistent across tasks tend to have higher magnitude}. We define an update direction as a sign of parameter change when trained on a given task, $\text{sgn}(\Delta\theta^p_t) = \text{sgn}(\theta^p_t - \theta^p_{t-1})$. For each parameter in each sequentially fine-tuned task vector, we calculate the number of consistent update directions $n$. Figure~\ref{fig:consistency-vs-magnitude} presents the relation of magnitude of task vectors' parameters and the consistency of update directions. We observe that the parameters with higher consistency tend to have higher magnitude. Therefore, we can think of maximum magnitude selection as a proxy for selecting the updates that multiple tasks agree~on.

\paragraph{\textbf{Contributions of task vectors.}}

\begin{figure}[t]
    \centering
    \includegraphics[width=0.98\linewidth]{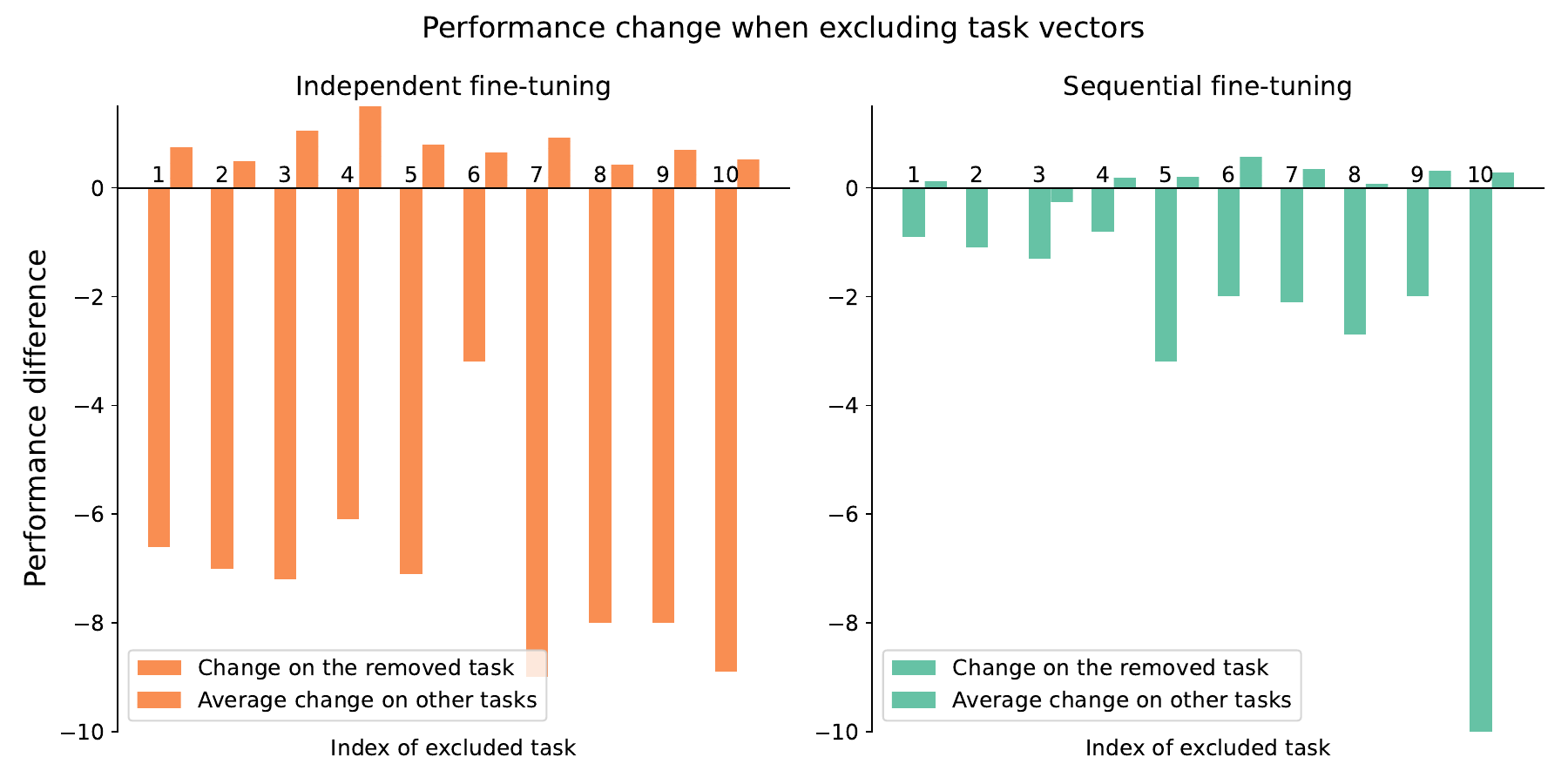}
    \caption{In an independent fine-tuning setting, the exclusion of a single task vector causes significant performance loss on the corresponding task. Models fine-tuned sequentially are much more robust to such an exclusion of non-last task vectors. It shows that the knowledge of previous tasks is partially retained in the later task vectors. We report the difference in accuracy (\%) between the model merged out of 10 task vectors and models merged out of 9 task vectors.}
    \label{fig:task-vector-exclusion}
\end{figure}

In this section, we present insights into the contributions of the particular task vectors to the final model. Firstly, we perform task vector exclusion experiments in the CIFAR100/10 setting. We merge 9 task vectors, excluding one of them, and compare it to the performance of 10 task vectors merged. We present the results in Figure~\ref{fig:task-vector-exclusion}. We observe that for independent
fine-tuning, removal of one task vector results in significant performance loss on the corresponding task. However, for sequentially fine-tuned models, the exclusion of a single task vector hurts the performance on the corresponding task much less. The only exception is the exclusion of the last task vector which uniquely contains the knowledge about the last task. This shows that the later task vectors retain some of the information about the previous tasks and the previous task vectors are less critical than when fine-tuning independently.

We demonstrate that this observation corresponds to the extent of contribution from the task vectors towards the merged model, which is quantified as the proportion of parameters chosen for the composite task vector. Figure~\ref{fig:task-vector-contribution} illustrates these contributions for the CIFAR100/10 experiment. When task-specific models are fine-tuned independently, their contributions are nearly evenly distributed. Yet, in the scenario where the model undergoes sequential fine-tuning, the contribution escalates with the task index, favoring models that have been fine-tuned across an increased number of tasks.

\paragraph{\textbf{Sensitivity to scaling factor.}}

\begin{figure}[t!]
    \centering
    \includegraphics[width=0.9\linewidth]{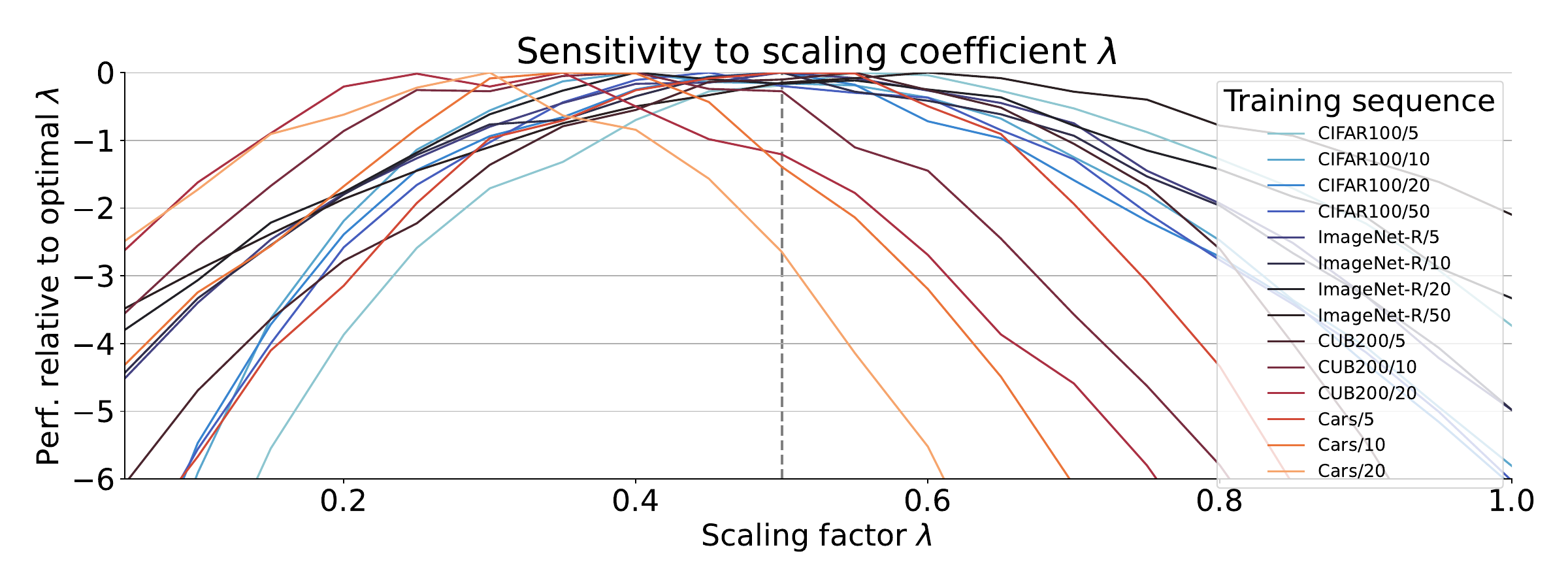}
    \caption{
        \methodshort{} is fairly stable across different scenarios when it comes to scaling coefficient $\lambda$. We report the accuracy (\%) relative to the accuracy with optimal $\lambda$.
    }
    \label{fig:lambda-sensitivity}
\end{figure}

Exemplar-free continual learning forbids storing data from previous tasks. Therefore we are not able to choose scaling factor $\lambda$ based on validation sets from all tasks as in~\cite{ilharco2023task} and we set a constant $\lambda=0.5$ for our method. Figure~\ref{fig:lambda-sensitivity} presents a sensitivity analysis of the scaling factor. We calculate the difference of the performance for $\lambda \in \{0.05, 0.1, \dots, 0.95, 1.0\}$ from the performance given an optimal $\lambda$. We observe that for 11 out of 14 scenarios, the results for selected $\lambda=0.5$ differ less than 0.5\% from the optimal selection. There are, however, several scenarios where selecting a better scaling coefficient would considerably improve the results. Note, that we only tuned $\lambda$ on CIFAR100/5 experiment and used the selected value across all other experiments (similar to other methods). 

\section{Extended analysis}

In this Section, we broaden the scope of our analysis. We investigate the impact of maximum magnitude selection merging on existing CL methods. We also study the impact of sequential fine-tuning on other model merging strategies in both CIL setting and on the popular eight datasets benchmark.

\paragraph{\textbf{Does model merging help CL methods?}}
In this section, we investigate if knowledge consolidation via model merging helps to improve the performance of CL methods. We modify \methodshort{} and instead of performing sequential fine-tuning, we train the model using one of the regularization-based CL methods. We present the results in Table~\ref{tab:CL-methods-with-merging}. We observe that adding model merging significantly improves the performance of LwF and EWC in almost every scenario. Interestingly, neither of these combinations significantly outperform \methodshort{} which uses naive sequential fine-tuning, traditionally known for causing catastrophic forgetting~\cite{french1999catastrophic, Mccloskey89}. These results show that model merging is a promising technique for consolidating the knowledge \textit{after} the training instead of \textit{during} the training.

\begin{table}[t]
\centering
\caption{
Knowledge consolidation step from \methodshort{} improves the performance of regularization-based CL methods. However, these combinations achieve an average performance on par with \methodshort{}. It suggests that forgetting mitigation techniques are less important when the knowledge is consolidated via model merging. 
}
\scalebox{0.72}{
\begin{tabular}{l|cccc|cccc|ccc|ccc|c}
\toprule
 & \multicolumn{4}{c|}{CIFAR100} & \multicolumn{4}{c|}{ImageNet-R} & \multicolumn{3}{c|}{CUB200} & \multicolumn{3}{c|}{Cars} & \multicolumn{1}{c}{Avg}\\
Method & /5 & /10 & /20 & /50 & /5 & /10 & /20 & /50 & /5 & /10 & /20 & /5 & /10 & /20 & \\
\midrule
LwF & 83.25 & 73.45 & 72.05 & 68.84 & 81.15 & 82.97 & 81.82 & 80.32 & 65.12 & 60.67 & 58.89 & 71.72 & 69.84 & 62.98 & 72.36 \\
LwF + \methodshort{} & 82.68 & 77.61 & 75.81 & 72.65 & 82.55 & 82.52 & 81.98 & 80.63 & 64.53 & 61.17 & 59.60 & 73.29 & 71.04 & 67.85 & 73.85 \\
$\Delta$ & \textcolor{Red}{-0.57} & \textcolor{Green}{+4.16} & \textcolor{Green}{+3.76} & \textcolor{Green}{+3.81} & \textcolor{Green}{+1.40} & \textcolor{Red}{-0.45} & \textcolor{Green}{+0.16} & \textcolor{Green}{+0.31} & \textcolor{Red}{-0.59} & \textcolor{Green}{+0.50} & \textcolor{Green}{+0.71} & \textcolor{Green}{+1.57} & \textcolor{Green}{+1.20} & \textcolor{Green}{+4.87} & \textcolor{Green}{+1.49} \\
\midrule
EWC & 84.41 & 76.24 & 75.39 & 72.97 & 82.15 & 82.42 & 81.48 & 81.47 & 59.10 & 54.49 & 53.31 & 69.46 & 60.78 & 57.42 & 70.79 \\
EWC + \methodshort{} & 82.34 & 77.73 & 77.66 & 77.03 & 82.07 & 83.02 & 82.35 & 81.60 & 63.57 & 60.61 & 59.15 & 72.83 & 69.59 & 66.00 & 73.97 \\
$\Delta$ & \textcolor{Red}{-2.07} & \textcolor{Green}{+1.49} & \textcolor{Green}{+2.27} & \textcolor{Green}{+4.06} & \textcolor{Red}{-0.08} & \textcolor{Green}{+0.60} & \textcolor{Green}{+0.87} & \textcolor{Green}{+0.13} & \textcolor{Green}{+4.47} & \textcolor{Green}{+6.12} & \textcolor{Green}{+5.84} & \textcolor{Green}{+3.37} & \textcolor{Green}{+8.81} & \textcolor{Green}{+8.58} & \textcolor{Green}{+3.18} \\
\midrule
\methodshort{} & 84.16 & 80.41 & 78.49 & 76.75 & 83.60 & 83.33 & 82.27 & 81.75 & 63.89 & 60.74 & 58.90 & 73.61 & 69.28 & 65.84 & 74.50 \\
\bottomrule
\end{tabular}
}
\label{tab:CL-methods-with-merging}
\end{table}

\paragraph{\textbf{Sequential fine-tuning improves various merging methods.}}
In this Section, we investigate how well sequential fine-tuning combines with different merging methods. Table~\ref{tab:ablation_seq_ft_CIL} presents the results of merging independent and sequential fine-tunings with different methods in class-incremental scenarios. We observe that all merging methods benefit from sequential fine-tuning in most of the scenarios, achieving from 1.3\% to 3.3\% better average results.
Table~\ref{tab:ablation_seq_ft_8datasets} presents the results of a similar experiment on eight dataset benchmark. We observe significant improvement (up to 12 p.p.) introduced by sequential fine-tuning. It shows that sequential fine-tuning can be beneficial even when the tasks are mostly dissimilar. Interestlingly, RandMix, Avg and TIES combined with Seq FT achieve very similar results, while \methodshort{} outperforms them by over 3 p.p.

\begin{table}[t]
\centering
\caption{
    Different merging methods combined with independent (Ind) and sequential (Seq) fine-tuning. RandMix and Avg benefit from sequential fine-tuning in most of the scenarios while TIES and \methodshort{} benefit in all of the evaluated scenarios. The best results are in \textbf{bold}.
}
\scalebox{0.73}{
\begin{tabular}{ll|cccc|cccc|ccc|ccc|c}
\toprule
 && \multicolumn{4}{c|}{CIFAR100} & \multicolumn{4}{c|}{ImageNet-R} & \multicolumn{3}{c|}{CUB200} & \multicolumn{3}{c|}{Cars} & Avg\\
Method & FT & /5 & /10 & /20 & /50 & /5 & /10 & /20 & /50 & /5 & /10 & /20 & /5 & /10 & /20 \\
\midrule
RandMix & Ind & 81.55 & 77.04 & 75.36 & 72.91 & 83.10 & 81.88 & 80.18 & 78.50 & 59.86 & 58.53 & 58.08 & 67.32 & 65.62 & 64.95 & 71.78\\
& Seq & 82.70 & 79.17 & 77.66 & 76.48 & 82.63 & 82.67 & 82.18 & 81.73 & 62.39 & 58.18 & 57.32 & 71.83 & 65.90 & 62.18 & 73.07 \\
& $\Delta$ & \textcolor{Green}{+1.15} & \textcolor{Green}{+2.13} & \textcolor{Green}{+2.30} & \textcolor{Green}{+3.57} & \textcolor{Red}{-0.47} & \textcolor{Green}{+0.79} & \textcolor{Green}{+2.00} & \textcolor{Green}{+3.23} & \textcolor{Green}{+2.53} & \textcolor{Red}{-0.35} & \textcolor{Red}{-0.76} & \textcolor{Green}{+4.51} & \textcolor{Green}{+0.28} & \textcolor{Red}{-2.77} & \textcolor{Green}{+1.29} \\
\midrule
Avg & Ind & 81.41 & 77.04 & 75.29 & 72.92 & 83.08 & 81.87 & 80.27 & 78.53 & 59.77 & 58.44 & 58.01 & 67.37 & 65.59 & 64.88 & 71.75 \\
& Seq & 82.68 & 79.12 & 77.60 & 76.46 & 82.60 & 82.60 & 82.18 & 81.65 & 62.50 & 58.20 & 57.34 & 71.88 & 65.94 & 62.01 & 73.05 \\
& $\Delta$ & \textcolor{Green}{+1.27} & \textcolor{Green}{+2.08} & \textcolor{Green}{+2.31} & \textcolor{Green}{+3.54} & \textcolor{Red}{-0.48} & \textcolor{Green}{+0.73} & \textcolor{Green}{+1.91} & \textcolor{Green}{+3.12} & \textcolor{Green}{+2.73} & \textcolor{Red}{-0.24} & \textcolor{Red}{-0.67} & \textcolor{Green}{+4.51} & \textcolor{Green}{+0.35} & \textcolor{Red}{-2.87} & \textcolor{Green}{+1.30} \\
\midrule
TIES & Ind & 81.72 & 77.23 & 74.66 & 73.76 & 83.08 & 82.27 & 80.83 & 79.57 & 60.94 & 58.22 & 56.97 & 70.45 & 64.90 & 61.17 & 71.84 \\
& Seq & 83.03 & 80.04 & 77.77 & 76.28 & 83.28 & 82.50 & \textbf{82.27} & 81.23 & 63.32 & \textbf{60.77} & \textbf{59.36} & \textbf{73.90} & \textbf{70.40} & \textbf{67.06} & 74.37\\
& $\Delta$ & \textcolor{Green}{+1.31} & \textcolor{Green}{+2.81} & \textcolor{Green}{+3.11} & \textcolor{Green}{+2.52} & \textcolor{Green}{+0.20} & \textcolor{Green}{+0.23} & \textcolor{Green}{+1.44} & \textcolor{Green}{+1.66} & \textcolor{Green}{+2.38} & \textcolor{Green}{+2.55} & \textcolor{Green}{+2.39} & \textcolor{Green}{+3.45} & \textcolor{Green}{+5.50} & \textcolor{Green}{+5.89} & \textcolor{Green}{+2.53} \\
\midrule
\methodshort{} & Ind & 81.95 & 76.75 & 74.39 & 73.04 & 83.03 & 82.33 & 80.92 & 79.33 & 60.15 & 58.01 & 56.59 & 67.36 & 63.55 & 58.95 & 71.17 \\
& Seq & \textbf{84.16} & \textbf{80.41} & \textbf{78.49} & \textbf{76.75} & \textbf{83.60} & \textbf{83.33} & \textbf{82.27} & \textbf{81.75} & \textbf{63.89} & 60.74 & 58.90 & 73.61 & 69.28 & 65.84 & \textbf{74.50} \\
& $\Delta$ & \textcolor{Green}{+2.21} & \textcolor{Green}{+3.66} & \textcolor{Green}{+4.10} & \textcolor{Green}{+3.71} & \textcolor{Green}{+0.57} & \textcolor{Green}{+1.00} & \textcolor{Green}{+1.35} & \textcolor{Green}{+2.42} & \textcolor{Green}{+3.74} & \textcolor{Green}{+2.73} & \textcolor{Green}{+2.31} & \textcolor{Green}{+6.25} & \textcolor{Green}{+5.73} & \textcolor{Green}{+6.89} & \textcolor{Green}{+3.33} \\
\bottomrule
\end{tabular}
}
\label{tab:ablation_seq_ft_CIL}
\end{table}

\begin{table}[t]
\centering
\caption{
Sequential fine-tuning leads to significant improvement over the independent fine-tuning even when tasks do not share many similarities. $\Delta$Avg indicates the average gain from using sequential fine-tuning over independent fine-tuning when merging models with different strategies in 8 datasets scenario.
}
\scalebox{0.77}{
\begin{tabular}{lc|cccccccc|cc}
\toprule
Method & \hspace{0.1in}FT\hspace{0.1in} & \hspace{0.05in}Cars\hspace{0.05in} & \hspace{0.07in}DTD\hspace{0.07in} & \hspace{0.03in}EuroSAT\hspace{0.03in} & \hspace{0.03in}GTSRB\hspace{0.03in} & \hspace{0.03in}MNIST\hspace{0.03in} & \hspace{0.01in}RESISC45\hspace{0.01in} & \hspace{0.03in}SUN397\hspace{0.03in} & \hspace{0.05in}SVHN\hspace{0.05in} & \hspace{0.1in}Avg\hspace{0.1in} & $\Delta$Avg \\
\midrule
RandMix & Ind & 69.82 & 50.27 & 83.11 & 59.26 & 94.51 & 75.83 & 68.52 & 74.24 & 71.95 \\
& Seq & 83.56 & 51.06 & 96.96 & 50.40 & 99.35 & 89.97 & 69.82 & 95.47 & 79.57 & \textcolor{Green}{+7.62}\\
\midrule
Avg & Ind & 57.11 & 45.53 & 75.00 & 66.78 & 98.78 & 62.48 & 46.07 & 88.60 & 67.54 \\
& Seq & 83.58 & 51.54 & 96.81 & 50.36 & 99.35 & 90.16 & 69.88 & 95.52 & 79.65 & \textcolor{Green}{+12.11}\\
\midrule
TIES & Ind & 71.76 & 55.53 & 88.33 & 67.61 & 98.91 & 78.67 & 66.09 & 89.81 & 77.09 \\
& Seq & 85.96 & 52.02 & 95.85 & 54.33 & 99.05 & 87.21 & 71.48 & 90.37 & 79.53 & \textcolor{Green}{+2.44} \\
\midrule
\methodshort{} & Ind & 73.14 & 52.02 & 83.15 & 56.33 & 96.80 & 79.54 & 71.42 & 80.69 & 74.14 \\ 
& Seq & 83.65 & 57.34 & 94.67 & 67.59 & 99.14 & 91.17 & 73.23 & 95.09 & 82.74 & \textcolor{Green}{+8.60}\\
\bottomrule
\end{tabular}
}
\label{tab:ablation_seq_ft_8datasets}
\end{table}

\paragraph{\textbf{Starting point for fine-tuning.}}

In this section, we investigate the relevance of the starting weights for fine-tuning when merging with \methodshort{}. When fine-tune the model on task $D_t$, we experiment with starting from $\theta_0$ (independent fine-tuning), $\theta_{t-1}$ (sequential fine-tuning) and $\theta_1$. The last option follows the intuition that the model fine-tuned on the first task is adapted to the particular domain or task, \eg bird species classification, and may serve as an appropriate starting point for future tasks that share some similarities. We present the results in Table~\ref{tab:starting-theta-CIL}. We observe that starting from $\theta_1$ usually hurts the final performance of the model compared to independent fine-tuning except for fine-grained scenarios with sufficiently big first tasks (CUB200/5 and Cars/5). However, both of these approaches underperform compared to the sequential fine-tuning highlighting the importance of knowledge transfer.

\begin{table}[h]
\centering
\caption{
Starting fine-tuning from the model adapted to a single task ($\theta_1$) does not improve the final performance compared to starting from pre-trained weights ($\theta_0$).
Training sequentially (starting from $\theta_{t-1}$) achieves the best results. 
}
\scalebox{0.77}{
\begin{tabular}{l|cccc|cccc|ccc|ccc|c}
\toprule
Initial $\theta$  & \multicolumn{4}{c|}{CIFAR100} & \multicolumn{4}{c|}{ImageNet-R} & \multicolumn{3}{c|}{CUB200} & \multicolumn{3}{c|}{Cars} & \multicolumn{1}{c}{Avg}\\
for $D_t$ & /5 & /10 & /20 & /50 & /5 & /10 & /20 & /50 & /5 & /10 & /20 & /5 & /10 & /20 & \\
\midrule
$\theta_0$ & \underline{81.95} & \underline{76.75} & \underline{74.39} & \underline{73.04} & \underline{83.03} & \underline{82.33} & \underline{80.92} & \underline{79.33} & 60.15 & \underline{58.01} & \underline{56.59} & 67.36 & \underline{63.55} & 58.95 & \underline{71.17} \\
$\theta_1$ & 81.42 & 75.59 & 70.64 & 69.27 & 82.43 & 81.37 & 79.70 & 78.53 & \underline{61.48} & 57.51 & 55.70 & \underline{70.45} & 61.66 & \underline{59.48} & 70.37 \\
$\theta_{t-1}$ & \textbf{84.16} & \textbf{80.41} & \textbf{78.49} & \textbf{76.75} & \textbf{83.60} & \textbf{83.33} & \textbf{82.27} & \textbf{81.75} & \textbf{63.89} & \textbf{60.74} & \textbf{58.90} & \textbf{73.61} & \textbf{69.28} & \textbf{65.84} & \textbf{74.50} \\
\bottomrule
\end{tabular}
}
\label{tab:starting-theta-CIL}
\end{table}

\section{Conclusion}

In this paper, we introduced \methodshort{}, a novel approach to continual learning that leverages model merging via maximum magnitude selection alongside sequential fine-tuning. Our findings underscore the potential of model merging as a viable solution to the challenges of continual learning. The synergy between sequential fine-tuning and maximum magnitude weight selection emerges as a pivotal factor in this success. It opens up possibilities for future research direction focused on developing fine-tuning methods that facilitate model merging or finding new, more effective strategies for selecting important parameters in realms of continual learning.

\section*{Acknowledgments}
Daniel Marczak is supported by National Centre of Science (NCN, Poland) Grant No. 2021/43/O/ST6/02482.
This research was partially funded by National Science Centre, Poland, grant no: 2020/39/B/ST6/01511, 2022/45/B/ST6/02817 and 2023/51/D/ST6/02846.
Bartłomiej Twardowski acknowledges the grant RYC2021-032765-I.
This paper has been supported by the Horizon Europe Programme (HORIZON-CL4-2022-HUMAN-02) under the project "ELIAS: European Lighthouse of AI for Sustainability", GA no. 101120237.
We gratefully acknowledge Polish high-performance computing infrastructure PLGrid (HPC Center: ACK Cyfronet AGH) for providing computer facilities and support within computational grant no. PLG/2023/016393.

\bibliographystyle{splncs04}
\bibliography{egbib}

\appendix
\clearpage
\setcounter{page}{1}

\section{More results}

\subsection{CIL results with different backbones}

We replicate our main results from Table~\ref{tab:results-CIL} with two different, stronger backbones: ViT-L-14 pre-trained on WebImageText~\cite{radford2021learning} (different architecture, the same pre-training dataset) in Table~\ref{tab:results-CIL-ViT-L-14} and ViT-B-16 pre-trained on LAION-400M (the same architecture, different pre-training dataset) in Table~\ref{tab:results-CIL-ViT-B-16-LAION400M}. \methodshort{} still outperforms both CL and merging-based baselines. We observe smaller improvement of \methodshort{} over the second best method than in Table~\ref{tab:results-CIL} as the room for improvement (defined as a difference in performance between joint and zero-shot model) is smaller for these stronger backbones.

\begin{table}
\centering
\caption{
Results with ViT-L-14 pre-trained on WebImageText~\cite{radford2021learning}.
}
\scalebox{0.75}{
\begin{tabular}{l|cccc|cccc|ccc|ccc|c}
\toprule
 & \multicolumn{4}{c|}{CIFAR100} & \multicolumn{4}{c|}{ImageNet-R} & \multicolumn{3}{c|}{CUB200} & \multicolumn{3}{c|}{Cars} & \multicolumn{1}{c}{Avg} \\
Method & /5 & /10 & /20 & /50 & /5 & /10 & /20 & /50 & /5 & /10 & /20 & /5 & /10 & /20 \\
\midrule
Zero-shot & \multicolumn{4}{c|}{75.82} & \multicolumn{4}{c|}{87.93} & \multicolumn{3}{c|}{62.86} & \multicolumn{3}{c|}{77.94} & 76.96 \\
Joint & \multicolumn{4}{c|}{93.49} & \multicolumn{4}{c|}{93.70} & \multicolumn{3}{c|}{87.00} & \multicolumn{3}{c|}{92.24} & 91.89 \\
\midrule
LwF & 86.89 & 83.21 & 82.12 & 82.23 & 90.92 & \textbf{91.80} & \textbf{91.28} & 90.60 & \textbf{73.96} & \textbf{71.09} & \underline{66.22} & 80.11 & \underline{79.14} & 75.92 & \underline{81.82} \\
EWC & \underline{88.72} & 85.41 & \underline{85.45} & \underline{83.34} & 91.13 & 91.37 & 91.08 & \textbf{90.98} & 68.66 & 66.45 & 64.65 & 78.97 & 74.05 & 69.74 & 80.71 \\
Rand Mix & 87.90 & \underline{85.60} & 83.67 & 81.36 & 91.48 & 90.45 & 89.32 & 88.20 & 67.74 & 65.31 & 64.34 & 80.08 & 77.96 & \underline{78.25} & 80.83 \\
Max Abs & 88.20 & 85.45 & 83.80 & 82.17 & \underline{91.67} & 91.15 & 90.05 & 88.95 & 67.45 & 65.33 & 63.39 & 79.48 & 75.91 & 75.46 & 80.60 \\
Avg & 87.87 & 85.54 & 83.69 & 81.36 & 91.47 & 90.42 & 89.32 & 88.20 & 67.60 & 65.29 & 64.36 & 80.14 & 77.91 & 78.26 & 80.82 \\
TIES & 88.35 & 85.40 & 84.05 & 82.46 & 91.58 & 90.75 & 89.90 & 88.77 & 67.86 & 65.48 & 64.45 & \underline{80.70} & 76.73 & 76.41 & 80.92 \\
\midrule
\methodshort{} & \textbf{88.90} & \textbf{87.24} & \textbf{86.18} & \textbf{84.97} & \textbf{91.95} & \underline{91.63} & \underline{91.22} & \underline{90.72} & \underline{70.30} & \underline{67.57} & \textbf{66.98} & \textbf{84.62} & \textbf{80.85} & \textbf{78.80} & \textbf{82.99} \\
\bottomrule
\end{tabular}
}
\label{tab:results-CIL-ViT-L-14}
\end{table}

\begin{table}
\centering
\caption{
Results with ViT-B-16 pre-trained on LAION-400M~\cite{schuhmann2021laion400m}.
}
\scalebox{0.75}{
\begin{tabular}{l|cccc|cccc|ccc|ccc|c}
\toprule
 & \multicolumn{4}{c|}{CIFAR100} & \multicolumn{4}{c|}{ImageNet-R} & \multicolumn{3}{c|}{CUB200} & \multicolumn{3}{c|}{Cars} & \multicolumn{1}{c}{Avg} \\
Method & /5 & /10 & /20 & /50 & /5 & /10 & /20 & /50 & /5 & /10 & /20 & /5 & /10 & /20 \\
\midrule
Zero-shot & \multicolumn{4}{c|}{71.29} & \multicolumn{4}{c|}{77.08} & \multicolumn{3}{c|}{64.64} & \multicolumn{3}{c|}{83.65} & 74.17 \\
Joint & \multicolumn{4}{c|}{90.79} & \multicolumn{4}{c|}{86.17} & \multicolumn{3}{c|}{81.31} & \multicolumn{3}{c|}{90.65} & 87.41 \\
\midrule
LwF & 84.14 & 78.09 & 75.77 & 75.13 & \underline{82.90} & \underline{82.80} & \underline{82.17} & \underline{80.82} & \textbf{71.92} & \textbf{67.93} & 65.07 & 81.12 & \underline{83.70} & 83.12 & \underline{78.19} \\
EWC & \underline{84.64} & \underline{80.22} & \underline{78.23} & \underline{75.85} & 82.02 & 81.52 & 81.32 & \textbf{80.93} & 65.83 & 64.88 & 62.74 & 82.88 & 76.99 & 76.36 & 76.74 \\
Rand Mix & 82.44 & 79.50 & 76.79 & 74.19 & 82.57 & 80.75 & 78.93 & 77.50 & 66.05 & 65.72 & \underline{65.34} & 84.59 & 83.39 & \textbf{83.58} & 77.24 \\
Max Abs & 82.89 & 79.57 & 77.05 & 75.07 & 82.75 & 81.58 & 79.90 & 78.02 & 65.77 & 64.50 & 64.01 & 84.58 & 81.89 & 80.81 & 77.03 \\
Avg & 82.41 & 79.53 & 76.80 & 74.16 & 82.57 & 80.78 & 78.92 & 77.48 & 66.10 & 65.74 & 65.31 & 84.67 & 83.47 & \underline{83.56} & 77.25 \\
TIES & 82.66 & 79.77 & 77.37 & 75.18 & 82.72 & 81.33 & 79.88 & 78.15 & 66.17 & 65.24 & 64.51 & \underline{84.79} & 82.28 & 82.09 & 77.30\\
\midrule
\methodshort{} & \textbf{84.85} & \textbf{81.67} & \textbf{80.31} & \textbf{78.20} & \textbf{83.47} & \textbf{83.07} & \textbf{82.23} & \underline{80.82} & \underline{69.14} & \underline{67.05} & \textbf{65.41} & \textbf{86.86} & \textbf{84.27} & 83.11 & \textbf{79.32} \\
\bottomrule
\end{tabular}
}
\label{tab:results-CIL-ViT-B-16-LAION400M}
\end{table}

\subsection{Sign conflicts}
\begin{figure}[t]
    \begin{subfigure}{.49\textwidth}
        \centering
        \includegraphics[width=0.98\linewidth]{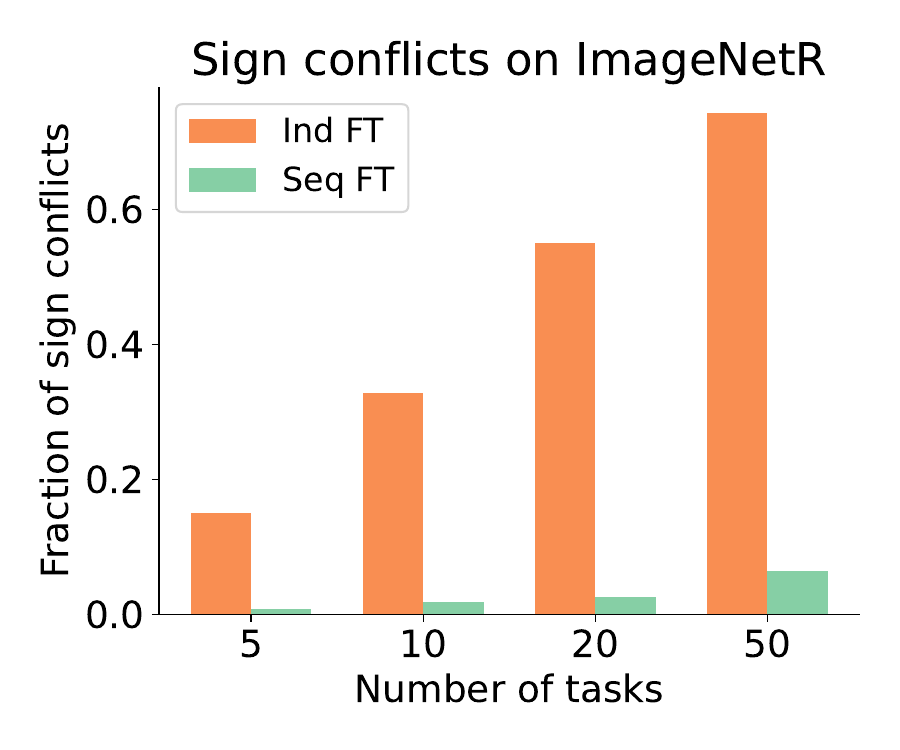}
    \end{subfigure}
    \begin{subfigure}{.49\textwidth}
        \centering
        \includegraphics[width=0.98\linewidth]{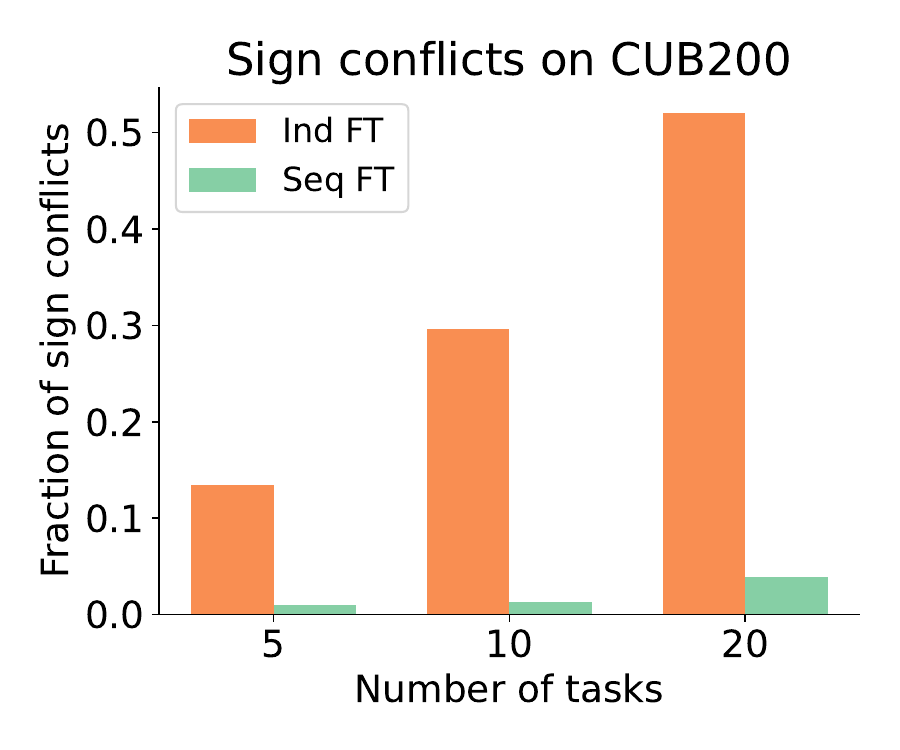}
    \end{subfigure}
    \begin{subfigure}{.49\textwidth}
        \centering
        \includegraphics[width=0.98\linewidth]{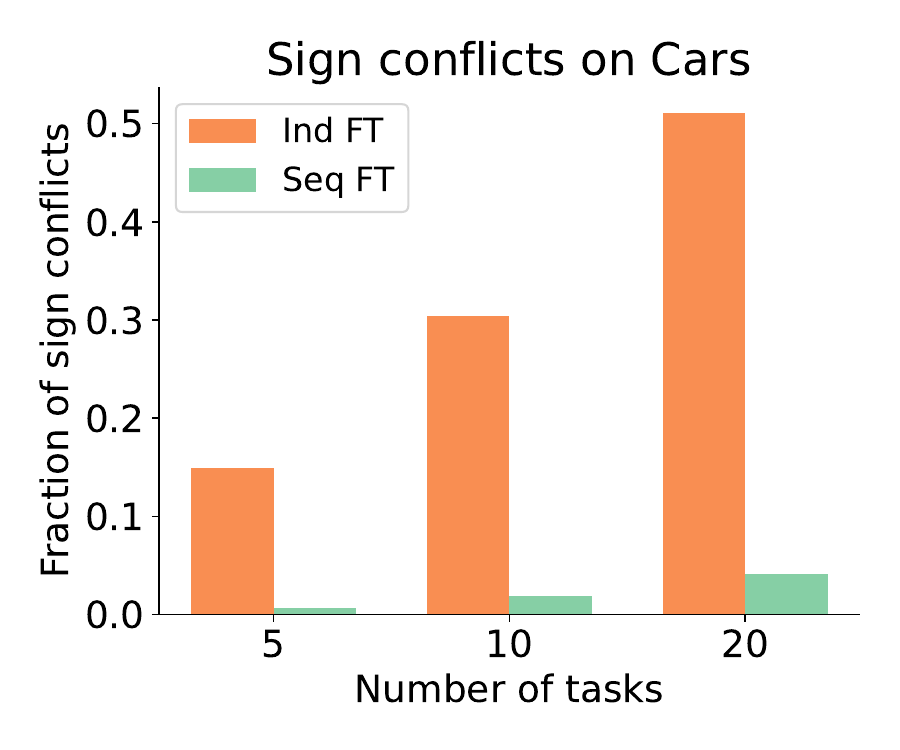}
    \end{subfigure}
    \begin{subfigure}{.49\textwidth}
        \centering
        \includegraphics[width=0.98\linewidth]{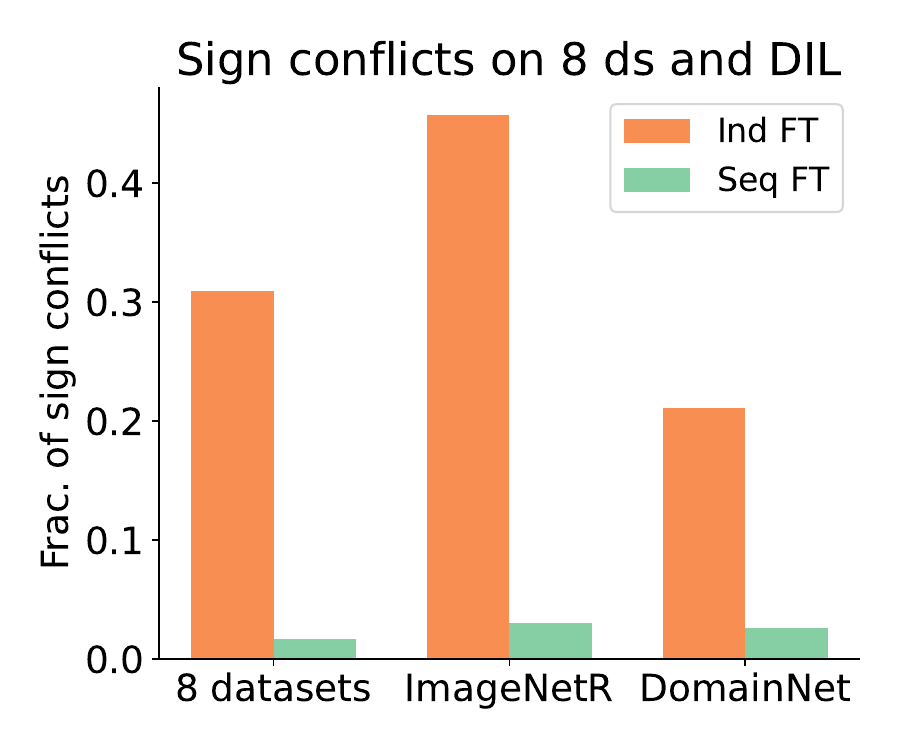}
    \end{subfigure}
    \caption{Sign conflicts for CIL, DIL and 8 datasets settings.}
    \label{fig:sign-conflicts-CIL-DIL-8ds}
\end{figure}

Fig.~\ref{fig:sign-conflicts-CIL-DIL-8ds} presents the sign conflicts for class-incremental,  domain-incremental and 8 datasets scenarios. We observe that sequential fine-tuning significantly reduces sign conflicts similarly to CIL results presented in Figure~3 in the main paper.

\subsection{Task agnostic per-task results}

\begin{figure}[t]
    \begin{subfigure}{.32\textwidth}
        \centering
        \includegraphics[width=0.999\linewidth]{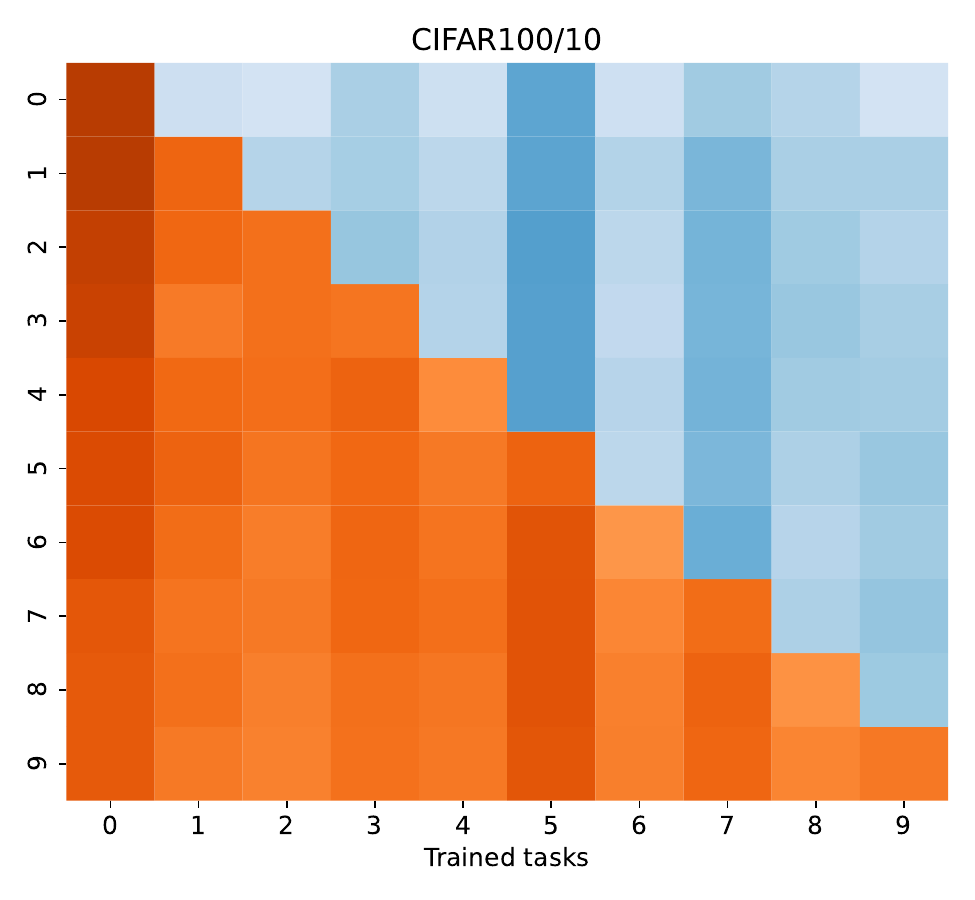}
    \end{subfigure}
    \begin{subfigure}{.32\textwidth}
        \centering
        \includegraphics[width=0.999\linewidth]{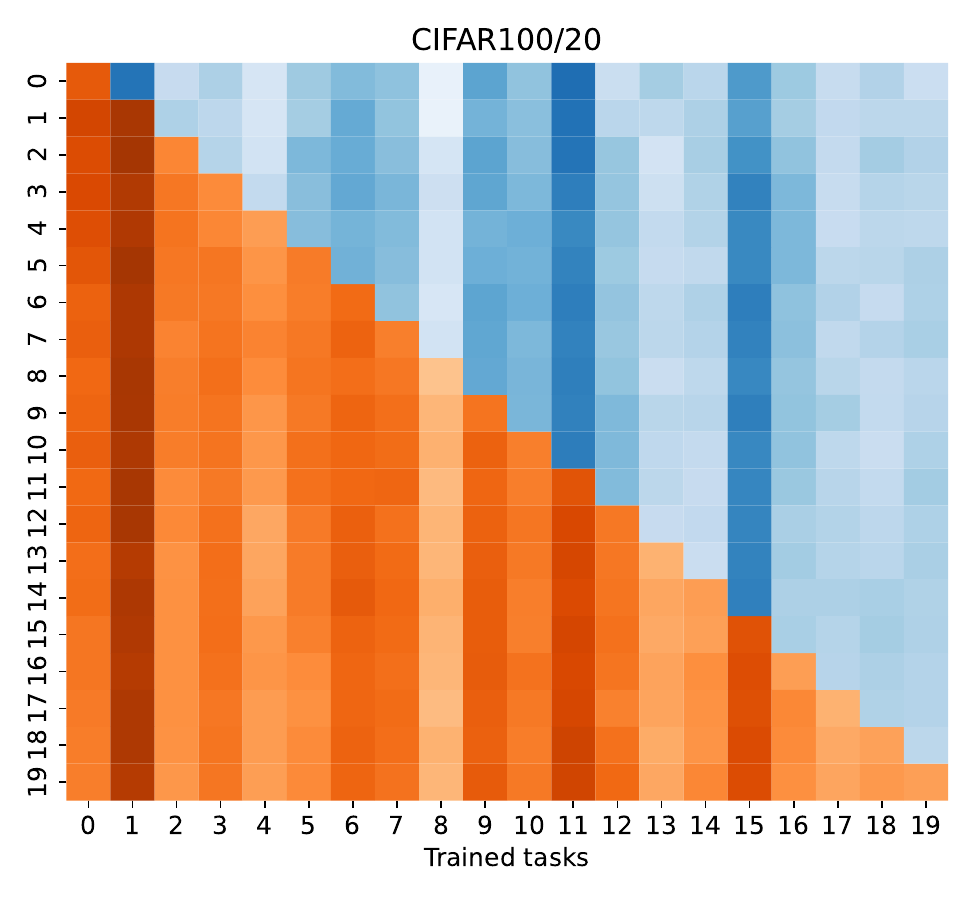}
    \end{subfigure}
    \begin{subfigure}{.32\textwidth}
        \centering
        \includegraphics[width=0.999\linewidth]{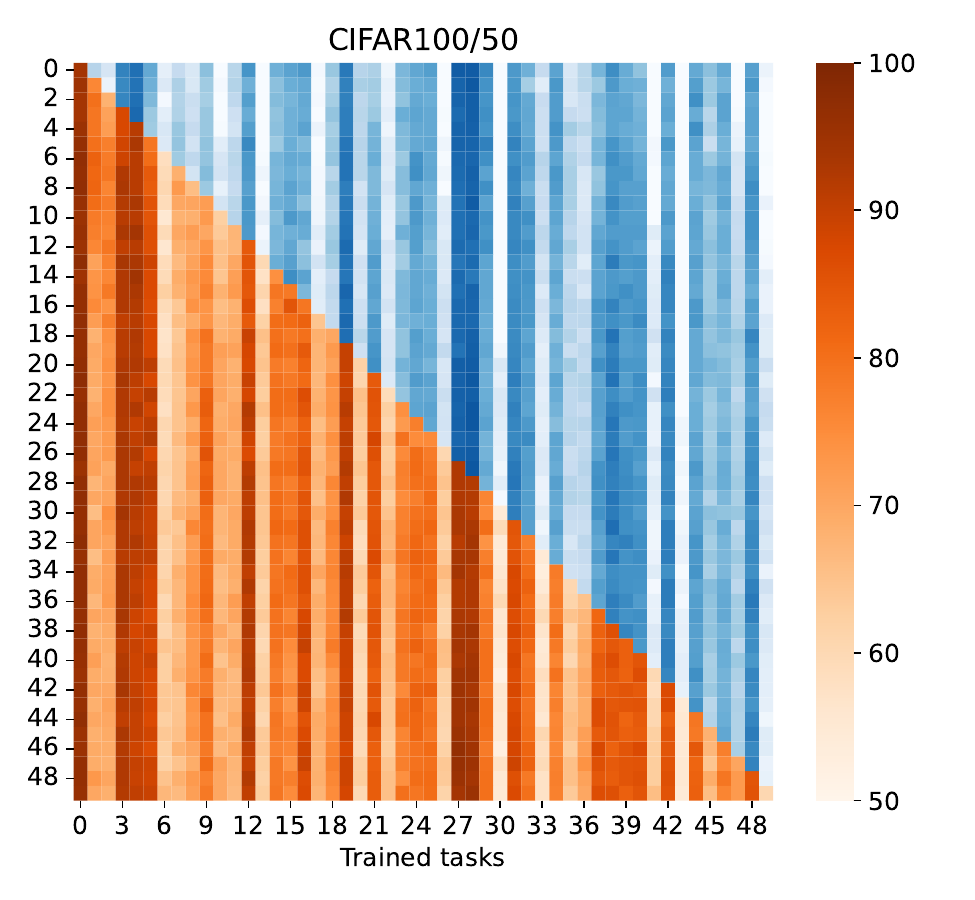}
    \end{subfigure}

    \begin{subfigure}{.32\textwidth}
        \centering
        \includegraphics[width=0.999\linewidth]{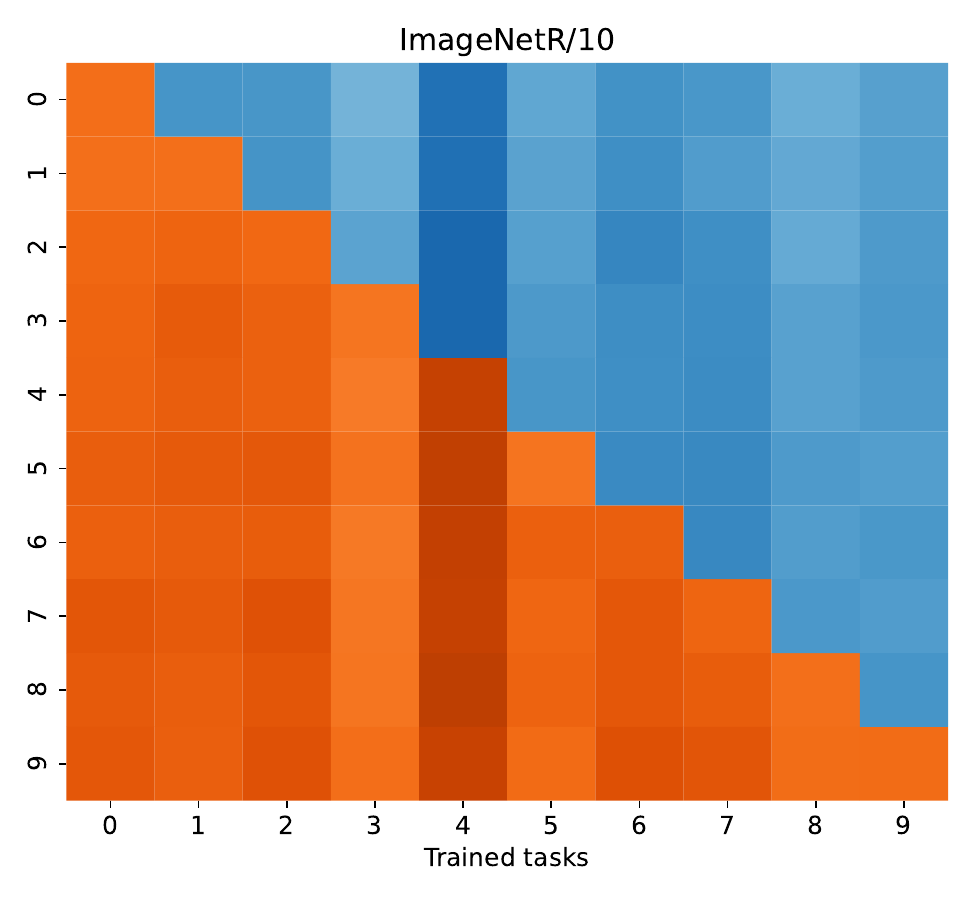}
    \end{subfigure}
    \begin{subfigure}{.32\textwidth}
        \centering
        \includegraphics[width=0.999\linewidth]{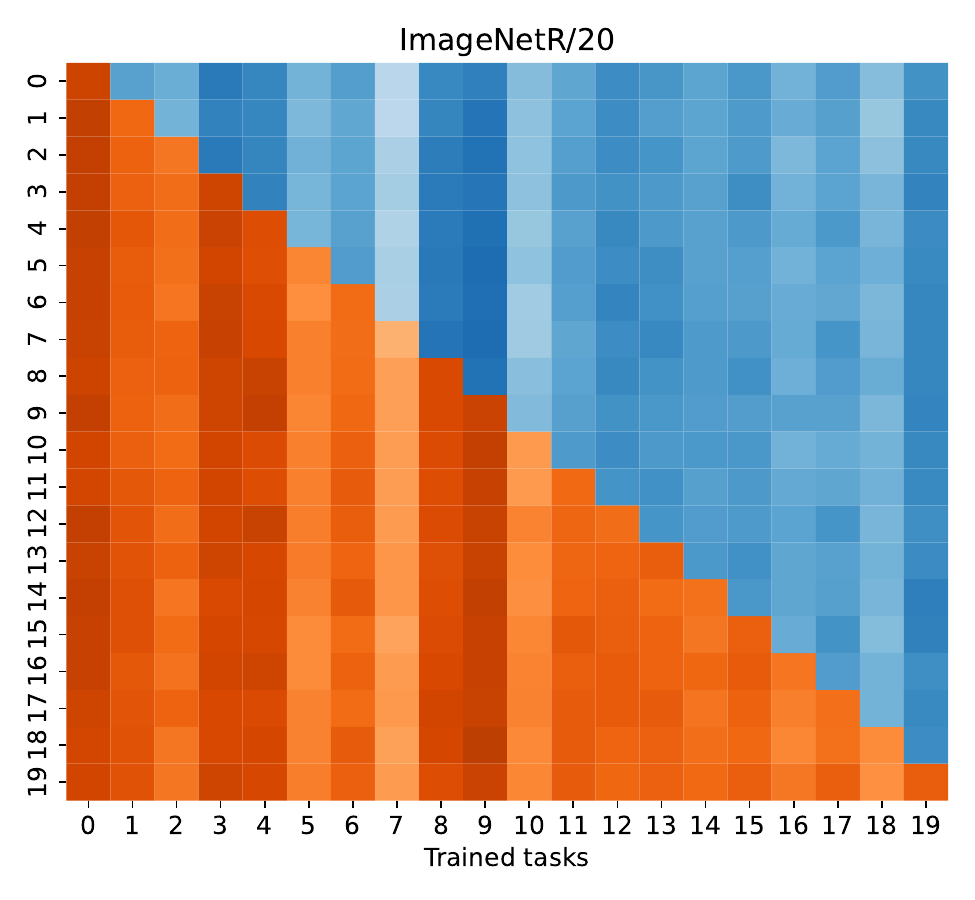}
    \end{subfigure}
    \begin{subfigure}{.32\textwidth}
        \centering
        \includegraphics[width=0.999\linewidth]{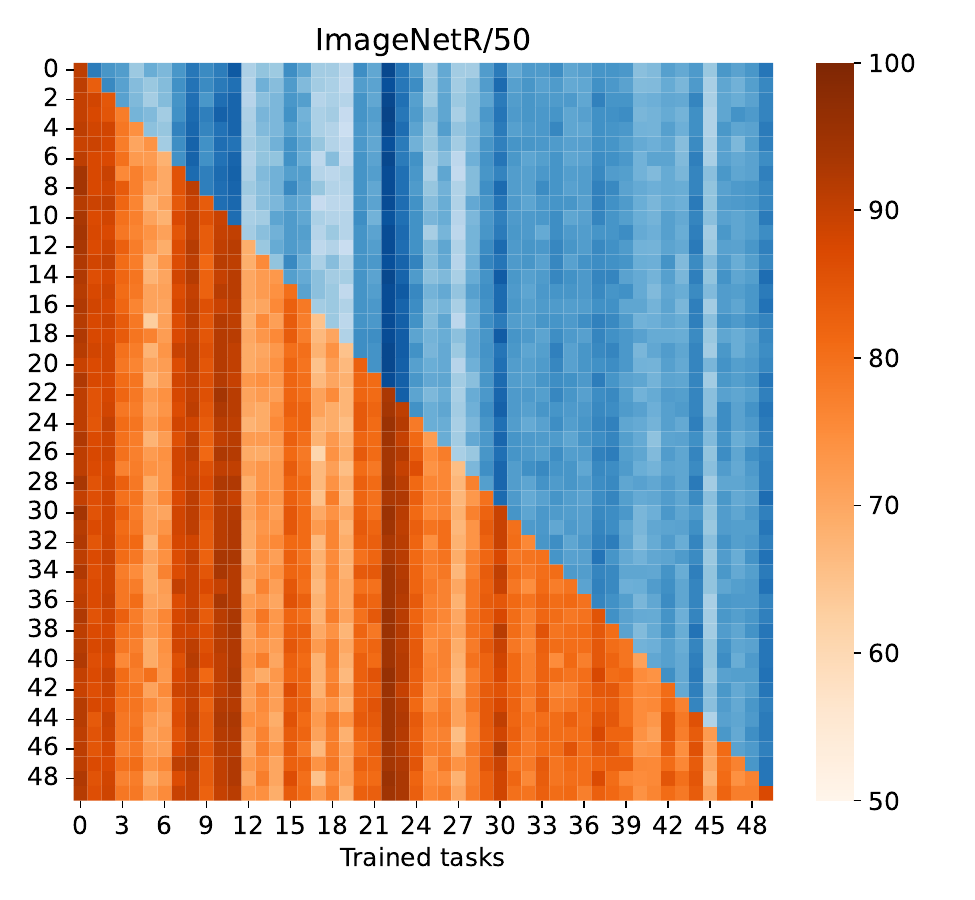}
    \end{subfigure}

    \begin{subfigure}{.32\textwidth}
        \centering
        \includegraphics[width=0.999\linewidth]{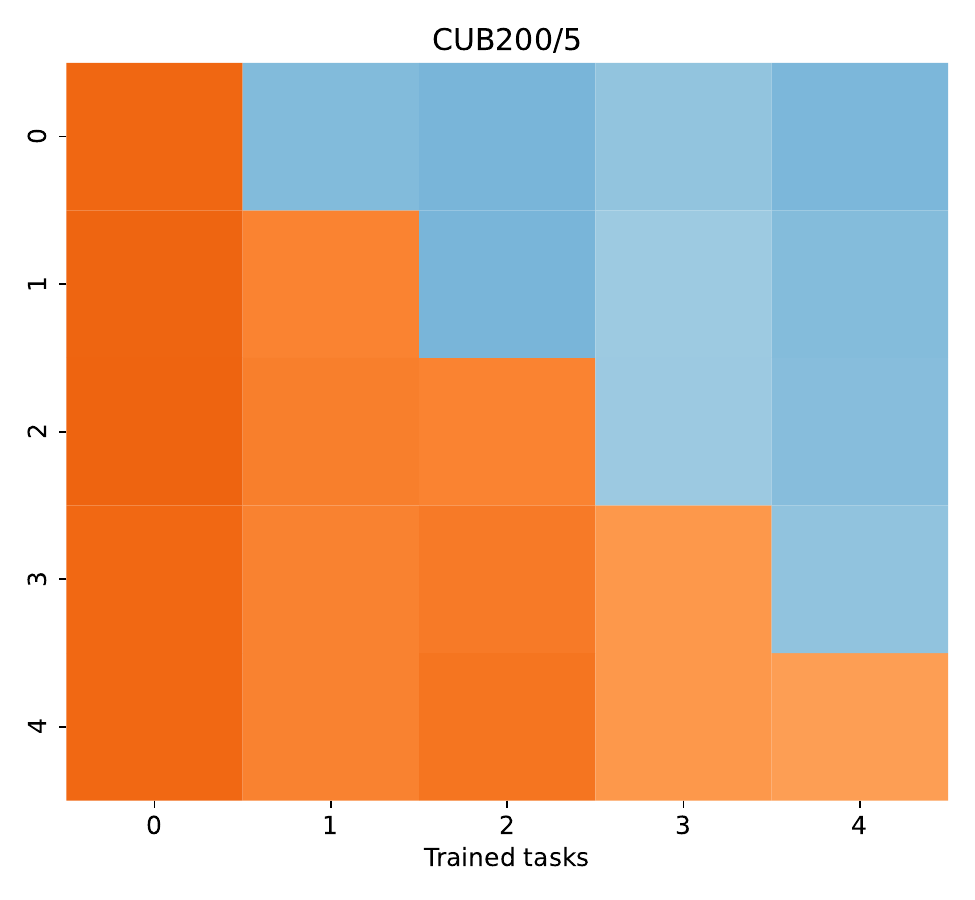}
    \end{subfigure}
    \begin{subfigure}{.32\textwidth}
        \centering
        \includegraphics[width=0.999\linewidth]{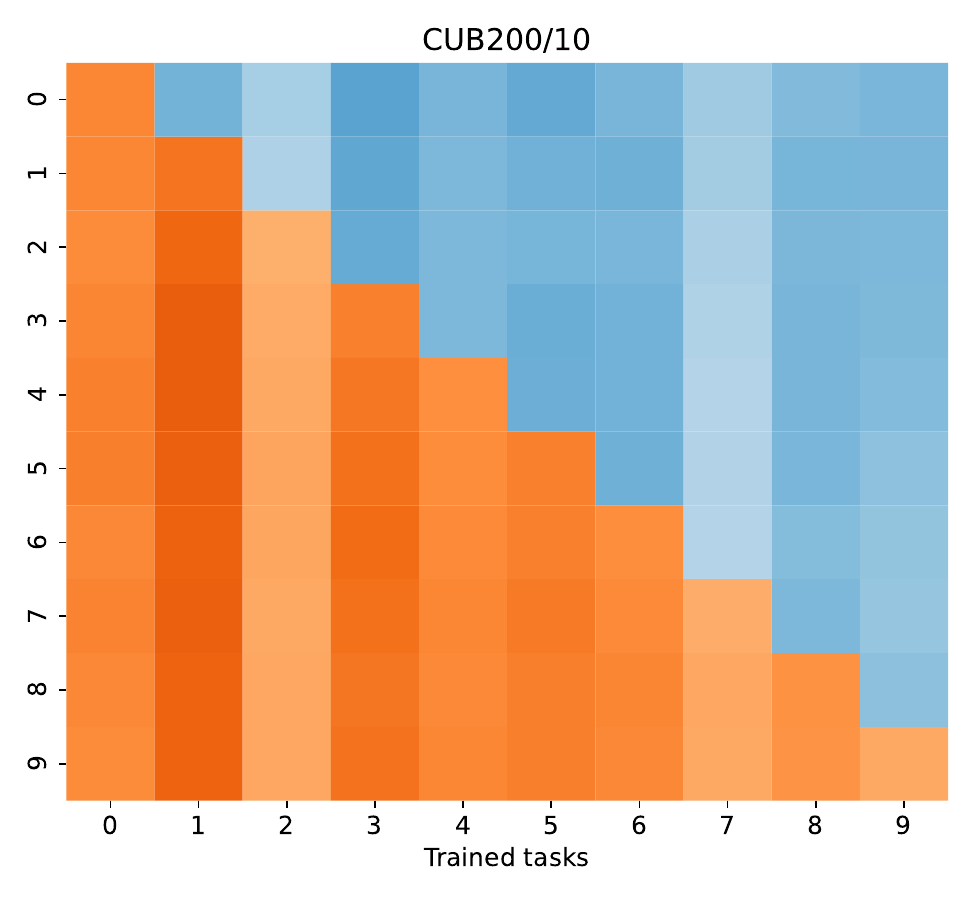}
    \end{subfigure}
    \begin{subfigure}{.32\textwidth}
        \centering
    \includegraphics[width=0.999\linewidth]{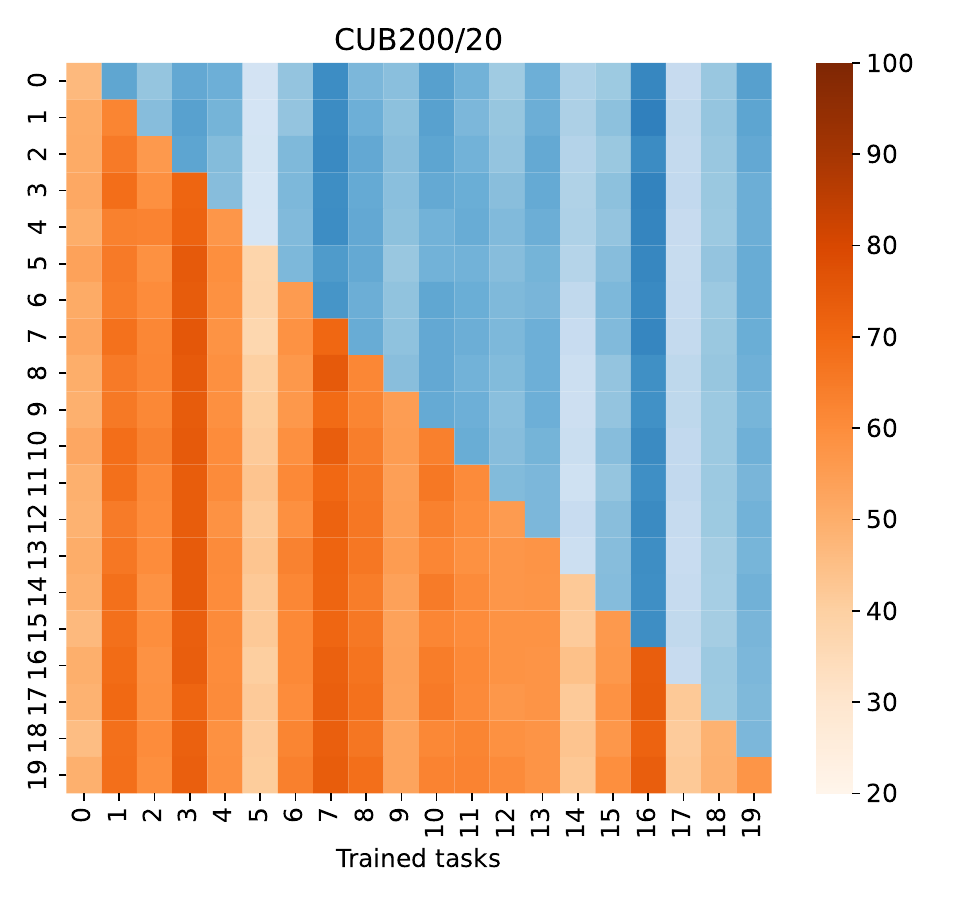}
    \end{subfigure}

    \begin{subfigure}{.32\textwidth}
        \centering
        \includegraphics[width=0.999\linewidth]{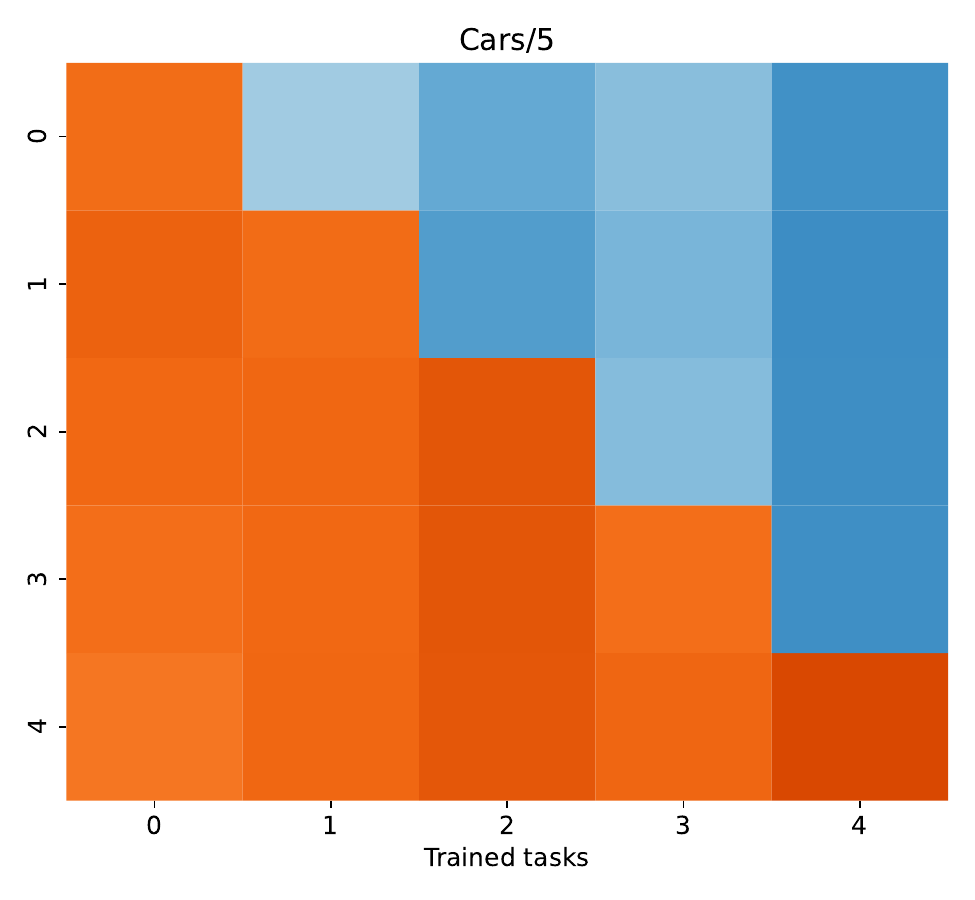}
    \end{subfigure}
    \begin{subfigure}{.32\textwidth}
        \centering
        \includegraphics[width=0.999\linewidth]{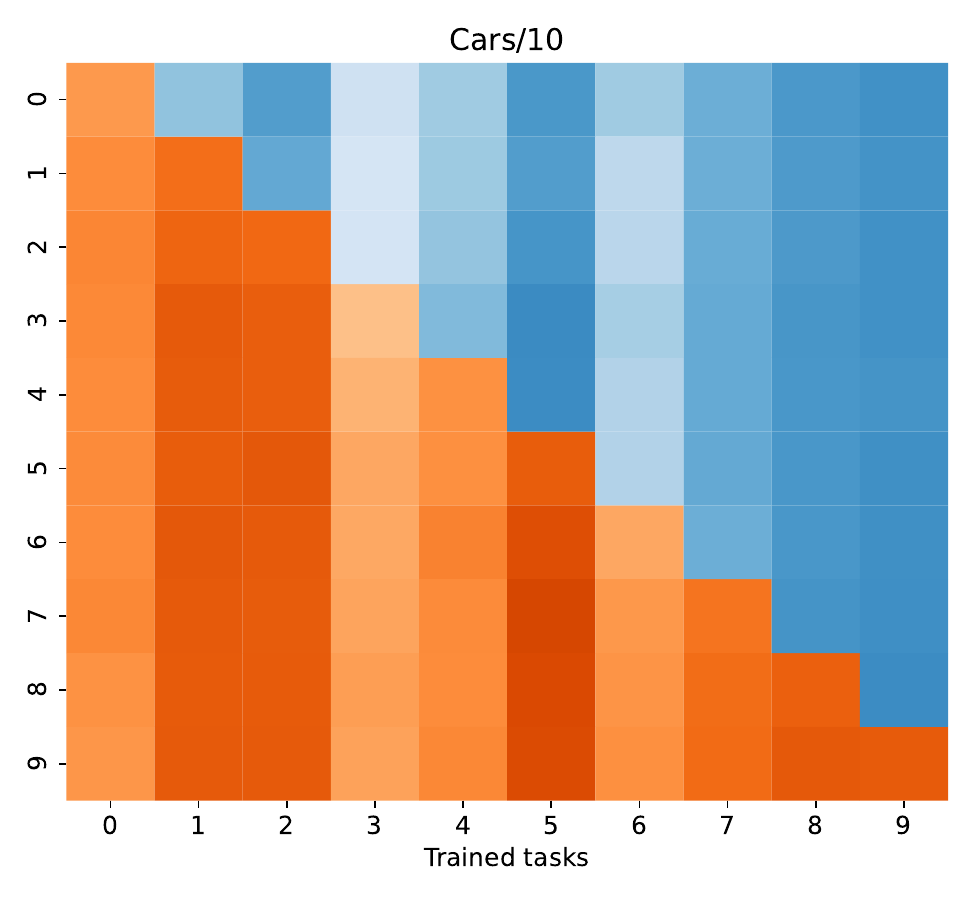}
    \end{subfigure}
    \begin{subfigure}{.32\textwidth}
        \centering
        \includegraphics[width=0.999\linewidth]{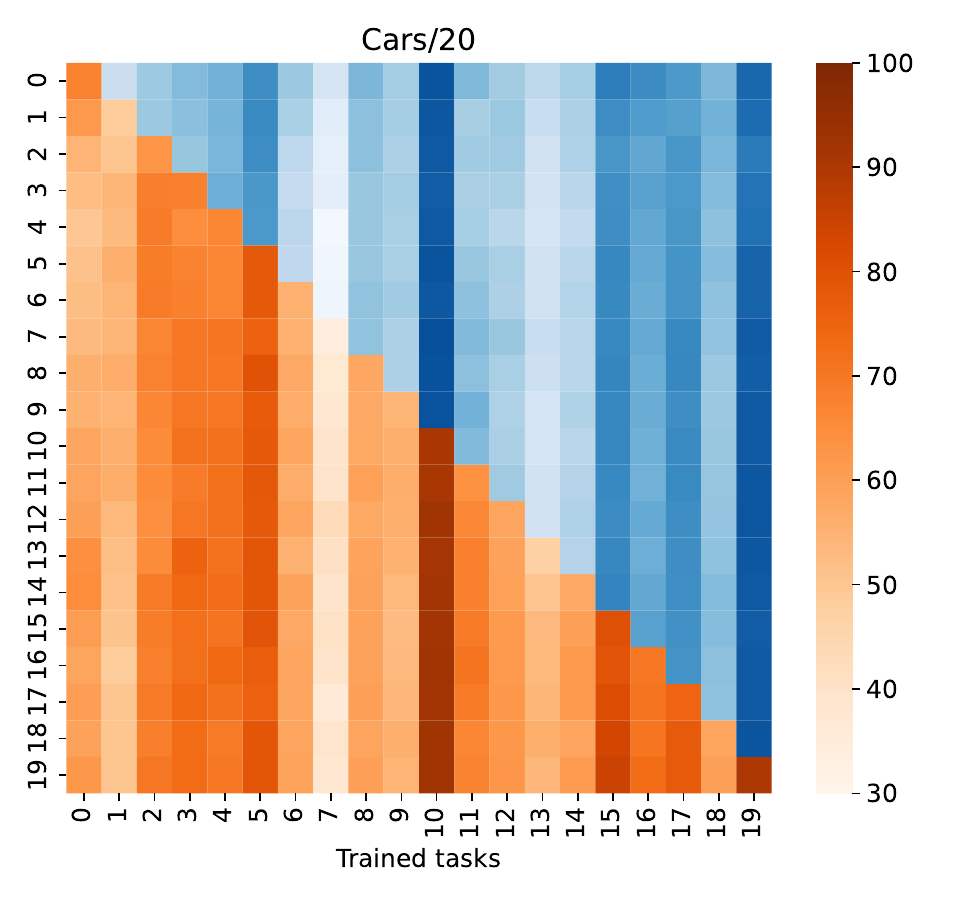}
    \end{subfigure}

    \caption{Task-agnostic results of \methodshort{} in different settings.}
    \label{fig:tag-results-more}
\end{figure}

Figure~\ref{fig:tag-results-more} presents more per-task task-agnostic results with \methodshort{}.

\section{Additional analyses}

\subsection{Layer-wise weight changes}
\begin{figure}[t]
    \begin{subfigure}{.49\textwidth}
        \centering
        \includegraphics[width=0.999\linewidth]{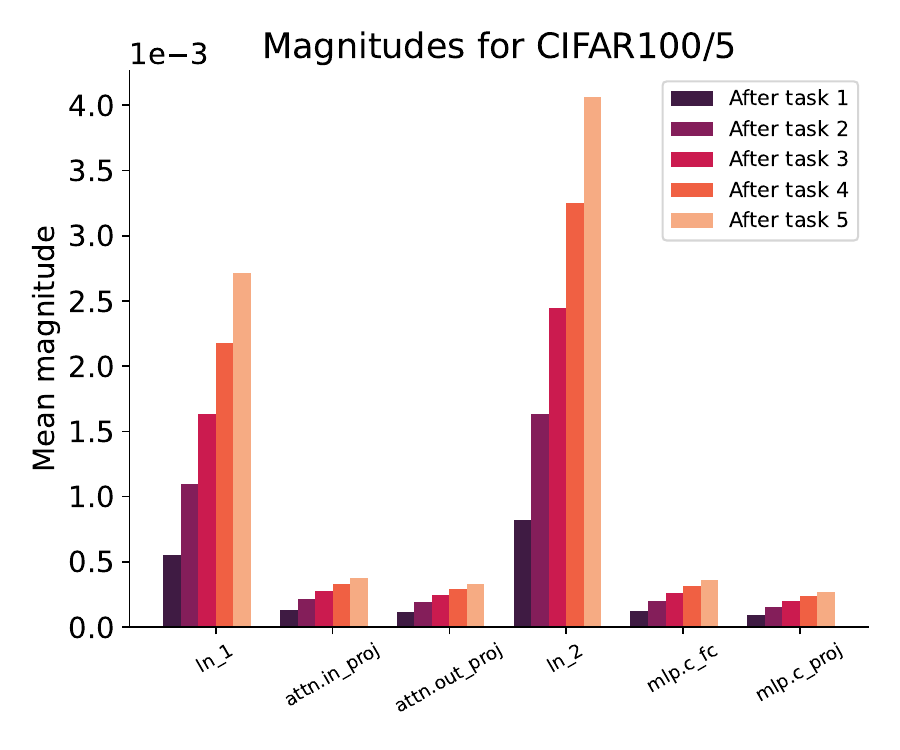}
    \end{subfigure}
    \begin{subfigure}{.49\textwidth}
        \centering
        \includegraphics[width=0.999\linewidth]{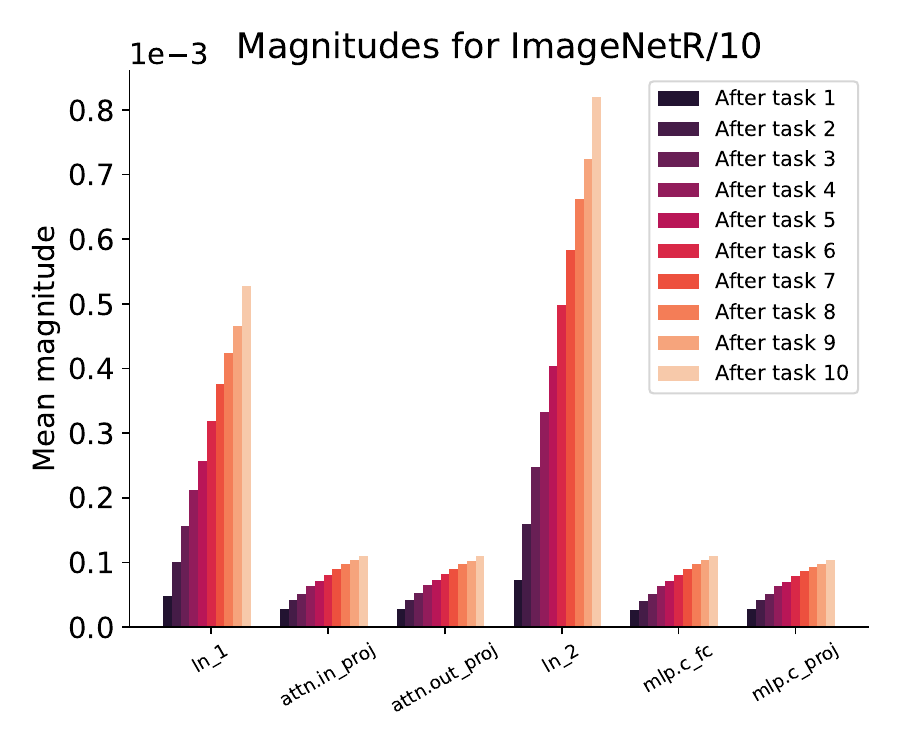}
    \end{subfigure}
    \begin{subfigure}{.49\textwidth}
        \centering
        \includegraphics[width=0.999\linewidth]{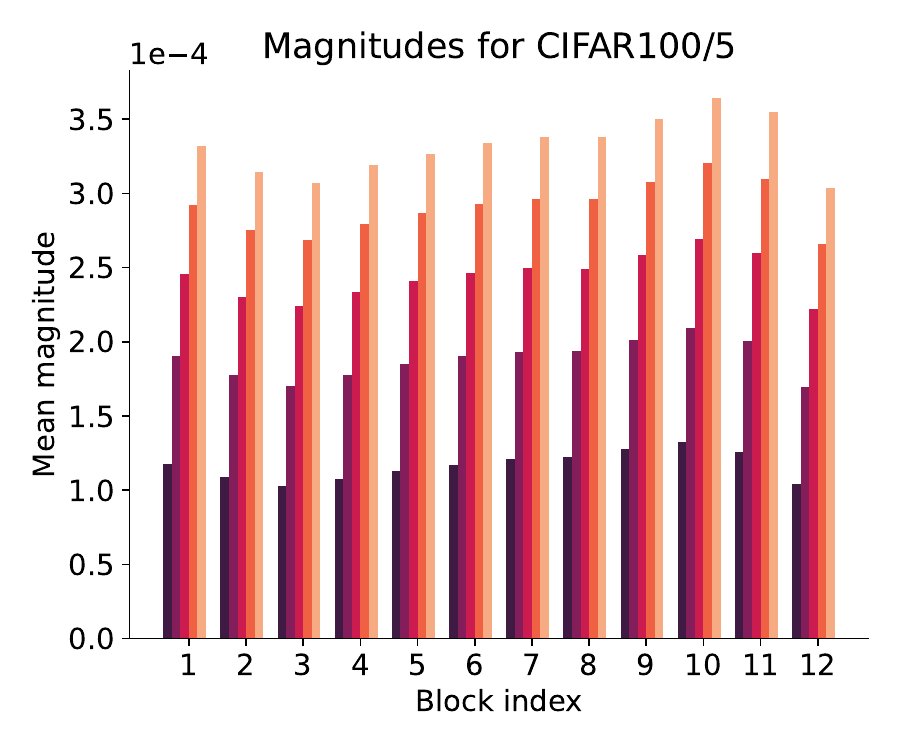}
    \end{subfigure}
    \begin{subfigure}{.49\textwidth}
        \centering
        \includegraphics[width=0.999\linewidth]{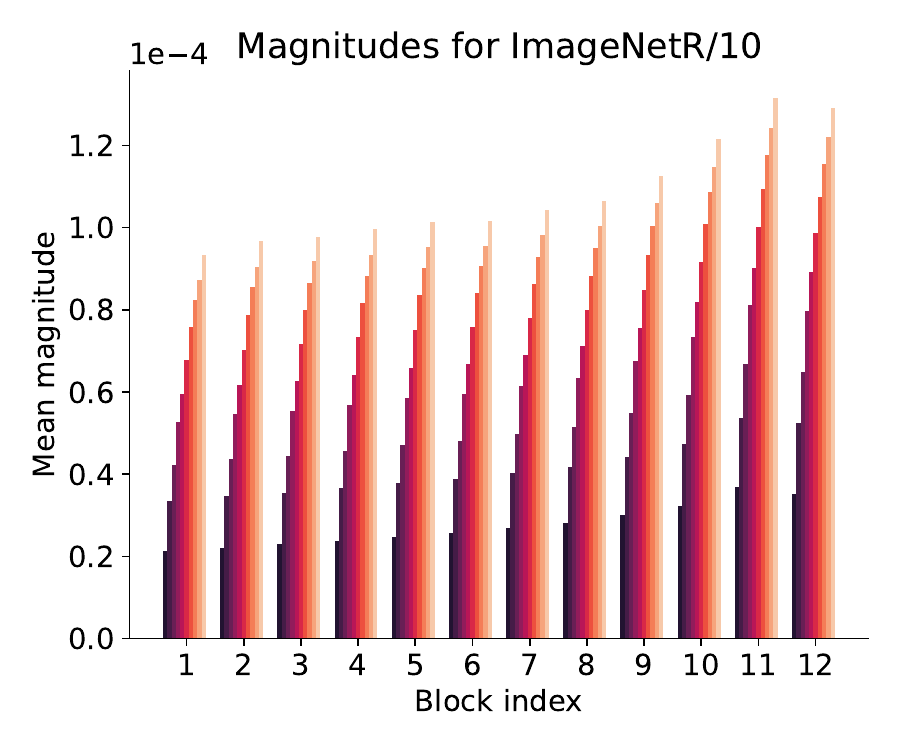}
    \end{subfigure}
    \caption{Mean magnitudes of $\tau_{\methodshort{}_t}$ parameters grouped by layer type (top) or block index (bottom) for CIFAR100/5 (left) and ImageNetR/10 (right). \textbf{Top:} parameters of LayerNorm layers change the most. \textbf{Bottom:} the magnitude of parameter change does not depend much on a block index (depth).}
    \label{fig:layerwise-magnitudes}
\end{figure}

To better undestand the process of fine-tuning and merging with \methodshort{}, we analyze the magnitudes of $\tau_{\methodshort{}_t}$ parameters. We group these parameters either by their type (e.g. layer normalization, attention or MLP) or by the block index to which they belong. We present the analysis in Figure~\ref{fig:layerwise-magnitudes} and observe that the magnitudes of layer normalization are much higher that the magnitudes of other layers. Moreover, magnitude seems not to depend on the depth. Note, that we only analyze weight matrices and disregard the biases. 

\subsection{Distribution of parameters in task vectors}

\begin{figure}[h!]
    \centering
    \includegraphics[width=0.999\linewidth]{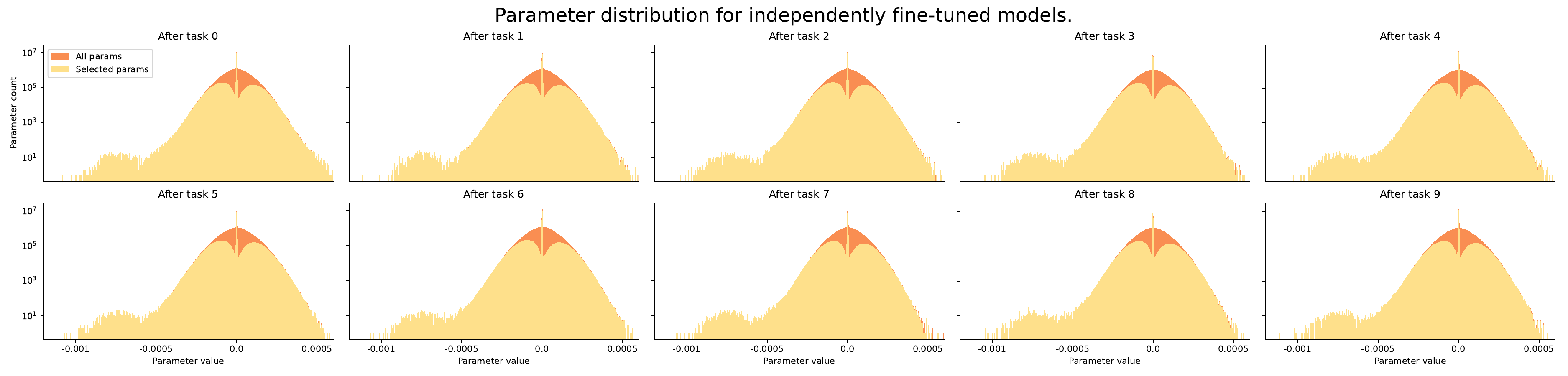}
    \includegraphics[width=0.999\linewidth]{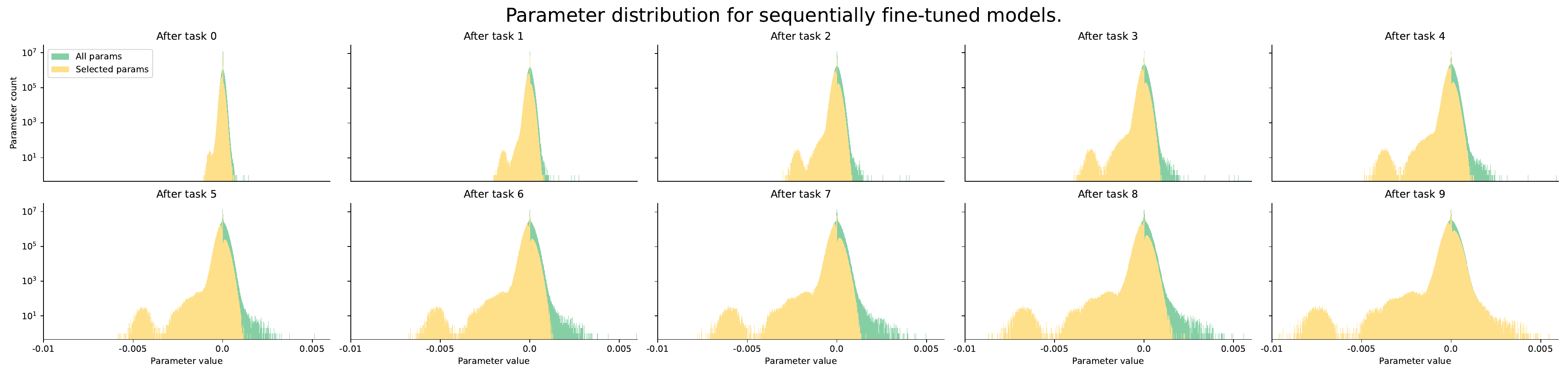}
    \caption{
        When fine-tuned independently (top), task vectors have similar distributions of parameters. Moreover, similar distribution contributes to the task vector merged by maximum magnitude selection.
        However, when fine-tuned sequentially (bottom), the distribution of parameters in task vectors differs -- later task vectors have larger parameters ans, as a results, they contribute more to the final task vector.
        Note that the vertical axis is logarithmic and that the scale of the independent and sequential distributions differ.
    }
    \label{fig:params-distr}
\end{figure}

Figure~\ref{fig:params-distr} presents the distribution of parameters in the task vectors. 

\end{document}